%% file: main.tex
\documentclass[10pt,twocolumn,letterpaper]{article}

\usepackage[pagenumbers]{cvpr} 
\usepackage{graphicx}
\usepackage{multirow}
\usepackage{amssymb}
\usepackage{amsmath}
\usepackage{wrapfig}
\usepackage{longtable}
\usepackage{supertabular}
\usepackage{textcomp}
\usepackage[accsupp]{axessibility}
\input{preamble}

\newcommand{\ours}{LLMVS}
\definecolor{cvprblue}{rgb}{0.21,0.49,0.74}
\usepackage[pagebackref,breaklinks,colorlinks,allcolors=cvprblue]{hyperref}
\usepackage{url}            
\usepackage{booktabs}       
\usepackage{amsfonts}       
\usepackage{nicefrac}       
\usepackage{microtype}      
\usepackage{xcolor}         
\usepackage{float}

\title{Video Summarization with Large Language Models}

\author{
Min Jung Lee\textsuperscript{1,2} \quad 
Dayoung Gong\textsuperscript{1} \quad 
Minsu Cho\textsuperscript{1} \quad \\ \\
\textsuperscript{1}Pohang University of Science and Technology (POSTECH) \quad\quad
\textsuperscript{2}GenGenAI
}

\begin{document}
\maketitle
\input{sec/0_abstract}    
\input{sec/1_introduction}
\input{sec/2_related_work}
\input{sec/3_method}
\input{sec/4_experiments}

\input{sec/5_conclusion}
\input{sec/6_ack}

{
    \small
    \bibliographystyle{ieeenat_fullname}
    \bibliography{main}
}
\input{sec/appendix}

\end{document}

%% file: preamble.tex
\usepackage[dvipsnames]{xcolor}
\usepackage{amsmath,amsfonts,bm}
\usepackage{colortbl}
\usepackage{float}

\definecolor{grey}{rgb}{0.9, 0.9, 0.9}
\newcommand{\rcol}{\rowcolor{grey}}
\usepackage{duckuments}
\usepackage{multicol}

\newcolumntype{C}[1]{>{\centering\let\newline\\\arraybackslash\hspace{0pt}}p{#1}}
\newcolumntype{L}[1]{>{\let\newline\\\arraybackslash\hspace{0pt}}p{#1}}

\usepackage{caption}

\usepackage{graphicx} 

%% file: sec/0_abstract.tex
\begin{abstract}
The exponential increase in video content poses significant challenges in terms of efficient navigation, search, and retrieval, thus requiring advanced video summarization techniques. Existing video summarization methods, which heavily rely on visual features and temporal dynamics, often fail to capture the semantics of video content, resulting in incomplete or incoherent summaries. To tackle the challenge, we propose a new video summarization framework that leverages the capabilities of recent Large Language Models (LLMs), expecting that the knowledge learned from massive data enables LLMs to evaluate video frames in a manner that better aligns with diverse semantics and human judgments, effectively addressing the inherent subjectivity in defining keyframes.
Our method, dubbed  \textbf{LLM}-based \textbf{V}ideo \textbf{S}ummarization (\textbf{\ours}), translates video frames into a sequence of captions using a Muti-modal Large Language Model (M-LLM) and then assesses the importance of each frame using an LLM, based on the captions in its local context. These local importance scores are refined through a global attention mechanism in the entire context of video captions, ensuring that our summaries effectively reflect both the details and the overarching narrative. Our experimental results demonstrate the superiority of the proposed method over existing ones in standard benchmarks, highlighting the potential of LLMs in the processing of multimedia content.

\end{abstract}

%% file: sec/1_introduction.tex
\input{figures/fig_teaser}
\section{Introduction}
Video summarization is essential in multimedia content processing, particularly as the exponential growth in video data has far exceeded human capacity for consumption. Every day, millions of videos are uploaded across platforms, posing significant challenges in efficient navigation, search, and retrieval of video content. 
Video summarization addresses these challenges by condensing lengthy videos into concise summaries that capture the essential content.
In response, researchers have explored automatic video summarization techniques aimed at producing videos that are shorter, more digestible, and appealing to users.
However, summarizing video content remains complex due to its varied nature and the subjective elements of effective summarization.

Previous video summarization methods~\cite{apostolidis2021combining, fajtl2019summarizing, ghauri2021supervised, jiang2022joint, zhu2020dsnet, son2024csta} have primarily focused on selecting important frames solely based on visual features. 
Recent multi-modal methods~\cite{narasimhan2021clip, he2023align, argaw2024scaling,li2023progressive} integrate both visual and language modalities to leverage the contextual richness of natural language.
However, these methods still prioritize visual features, incorporating textual data via an attention mechanism~\cite{vaswani2017attention}, where visual features serve as queries and language features as keys and values. 
While textual data helps to enhance the visual features, the main focus of video summarization still remains on visual content. 

The advent of Large Language Models (LLMs)~\cite{achiam2023gpt, touvron2023llama, touvron2023llama2, almazrouei2023falcon, anil2023palm} presents new opportunities for video summarization. LLMs have shown strong capabilities in contextual understanding~\cite{brown2020language, khashabi2020unifiedqa}, cross-domain reasoning~\cite{wei2022chain, kojima2022large}, and multimodal processing~\cite{liu2024visual, alayrac2022flamingo, li2022blip, chen2022visualgpt, driess2023palm}, allowing them to identify key moments based on semantic insights rather than visual saliency alone. 
Leveraging these strengths, we introduce \ours, an LLM-based video summarization framework that utilizes LLMs as important frame selectors, guided by textual data and embedded knowledge.

To this end, we propose a local-to-global video summarization model, as illustrated in Figure~\ref{fig:teaser}. 
First, we obtain textual data for each frame by generating textual descriptions from video frames using a pre-trained multi-modal LLM (M-LLM)~\cite{liu2024visual}.
Textual descriptions of video frames within a local window are fed into the LLM~\cite{touvron2023llama2}, along with structured instructions and examples in natural language, to perform in-context learning for video summarization. The LLM then evaluates the importance score of the center frame within the local context.
Unlike existing methods that rely on the end output of LLMs~\cite{zhao2023antgpt, khandelwal2023analyzing, saad2023pdftriage, zanella2024harnessing}, our method extracts the output embeddings from LLMs and apply self-attention on them to aggregate global context from the videos and make the final predictions. 
During learning, the M-LLM and LLM are frozen to preserve their general domain knowledge, and only the self-attention blocks are trained. 

Our contributions can be summarized as follows: 
\textbf{1)} We introduce \ours, a novel video summarization framework that leverages LLMs to utilize textual data and general knowledge in video summarization effectively.
\textbf{2)} The proposed local-to-global video summarization framework integrates local context via window-based aggregation and global context through self-attention, enabling a comprehensive understanding of video content.
\textbf{3)} Experimental results show that using output embeddings from LLMs is more effective for video summarization than using direct answers generated by LLMs.
\textbf{4)} Comprehensive results demonstrate the effectiveness of the proposed method, achieving state-of-the-art performance on the SumMe and TVSum datasets.

%% file: figures/fig_teaser.tex
\begin{figure}[h]
    \centering
    \scalebox{0.4}{
    \includegraphics{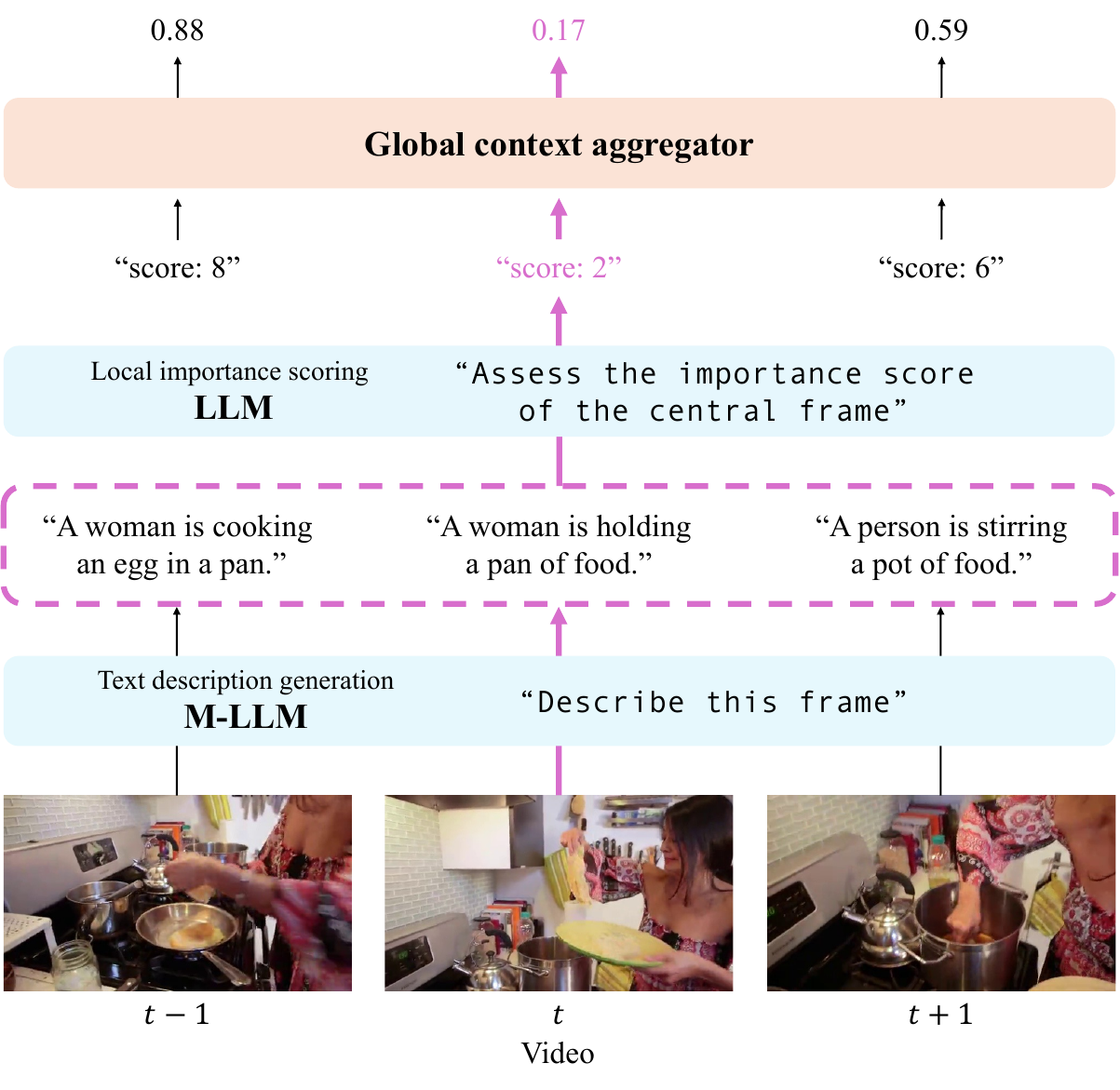}
    }
    \vspace{-0.3cm} 
    \caption{ \textbf{Video summarization with (M-)LLMs.}
    Given the input video frames, captions for each frame are generated using M-LLM. 
    For each frame at time-step $t$, the generated captions within a local window are grouped and provided as input to the LLM. The LLM is prompted to assess the importance score of the frame at time step $t$ by considering this local context. Finally, a global context aggregator produces the final predictions by taking into account the overall context of the entire video.
    Note that, in this illustration, the local window size is set to 3. 
}
    \label{fig:teaser}
    \vspace{-0.2cm} 
\end{figure}

%% file: sec/2_related_work.tex
\input{figures/fig_pipeline}

\section{Related Work}

\subsection{Video Summarization}
Recent advancements in video summarization have significantly leveraged deep learning techniques by capturing temporal dynamics. A notable direction in this domain employs LSTMs~\cite{zhang2016video, zhao2020tth, zhao2018hsa, wang2019stacked, ji2019video, yuan2019cycle, zhao2021reconstructive, hussain2019cloud} which are adept at capturing both short- and long-range dependencies in sequential frames.
A pioneering work by Zhang et al.\cite{zhang2016video} utilizes Long Short-Term Memory (LSTM) networks, leveraging their ability to model variable temporal dependencies among video frames. Building on this foundation, subsequent studies explored various LSTM-based architectures for video summarization, such as hierarchical frameworks\cite{zhao2018hsa}, stacked LSTMs~\cite{wang2019stacked}, and encoder-decoder structures~\cite{ji2019video}.
Transitioning to the utilization of self-attention mechanisms~\cite{apostolidis2021combining, fajtl2019summarizing, ghauri2021supervised, jiang2022joint, zhu2020dsnet, wang2020learning, xu2021cross}, VASNet~\cite{fajtl2019summarizing} employs soft self-attention. Other approaches introduce localization components to guide attention, such as DSNet~\cite{zhu2020dsnet}, which predicts the spatial offsets of interest regions, and iPTNet~\cite{jiang2022joint}, which integrates moment localization through collaborative learning. Positional encoding has also been explored, as in PGL-SUM~\cite{apostolidis2021combining}, which incorporates absolute position information into multi-head attention.
CSTA~\cite{son2024csta} initially extracts and concatenates frame features, representing the temporal sequence as an image. This representation is then processed by a 2D CNN, yielding attention maps that capture spatiotemporal dependencies. These models primarily rely on visual cues and temporal features. In contrast, our work leverages the capabilities of LLMs to incorporate semantic information, enriching the contextual understanding of the video.

\subsection{Multi-Modal Video Summarization}
Unlike unimodal methods that rely solely on visual frames, multimodal video summarization~\cite{he2023align, narasimhan2021clip, argaw2024scaling, li2023progressive} integrates multiple modalities, such as visual and textual features, to produce more comprehensive summaries.
CLIP-It~\cite{narasimhan2021clip} utilizes a cross-attention module between visual and textual features, both extracted using CLIP~\cite{radford2021learning}, allowing summarization to be conditioned on natural language. A2Summ~\cite{he2023align} introduces an alignment-guided self-attention module that effectively fuses different modalities by leveraging the temporal correspondence between video and text, incorporating captions generated by GPT-2~\cite{radford2019language} or existing transcript. Argaw et al.\cite{argaw2024scaling} propose a cross-modal attention mechanism to integrate multimodal cues from contextualized features, employing SRoBERTa-NLI-large\cite{reimers2019sentence} for sentence embedding and CLIP~\cite{radford2021learning} for visual features.
Prior multimodal video summarization methods~\cite{narasimhan2021clip, argaw2024scaling, he2023align} employ cross-attention mechanisms, where visual features act as queries and language features serve as keys and values. While these approaches incorporate language to enhance semantic understanding, they primarily focus on refining visual representations, often treating textual information as auxiliary information.
In this paper, \ours~leverages contextual understanding capabilities of LLMs for video summarization by utilizing both textual data and the general knowledge encoded in LLMs.

\subsection{Video Understanding with LLM}
Recent advancements in natural language processing (NLP) have been significantly driven by Large Language Models (LLMs)\cite{touvron2023llama, touvron2023llama2, achiam2023gpt, taori2023alpaca, team2023gemini, chiang2023vicuna}. The widespread adoption of these models spur the development of multimodal models that seamlessly integrate vision and text data\cite{zhang2023video, song2023moviechat, khandelwal2023analyzing, min2024morevqa, zhao2023antgpt}. MovieChat~\cite{song2023moviechat} enhances video understanding by processing video representations with a Q-former~\cite{li2023blip} and a linear projection layer. These components convert visual features into text space before feeding them into a LLM for user interaction. In the realm of video question answering, MoReVQA~\cite{min2024morevqa} employs LLM in a multistage modular reasoning framework that breaks down complex queries into event parsing, grounding, and reasoning stages to interpret complex queries. Similarly, AntGPT~\cite{zhao2023antgpt} addresses action anticipation task by leveraging LLMs to infer pseudo-ground truth from observed actions and generate future steps. These works highlight the versatility of M-LLMs in merging data modalities and transforming interactions across domains. Inspired by these advancements, our approach applies M-LLM and LLM to the video summarization task, leveraging their ability to incorporate semantic information and provide a richer contextual understanding of video content.

%% file: figures/fig_pipeline.tex
\begin{figure*}[h]
    \centering
    \includegraphics[width=\textwidth] {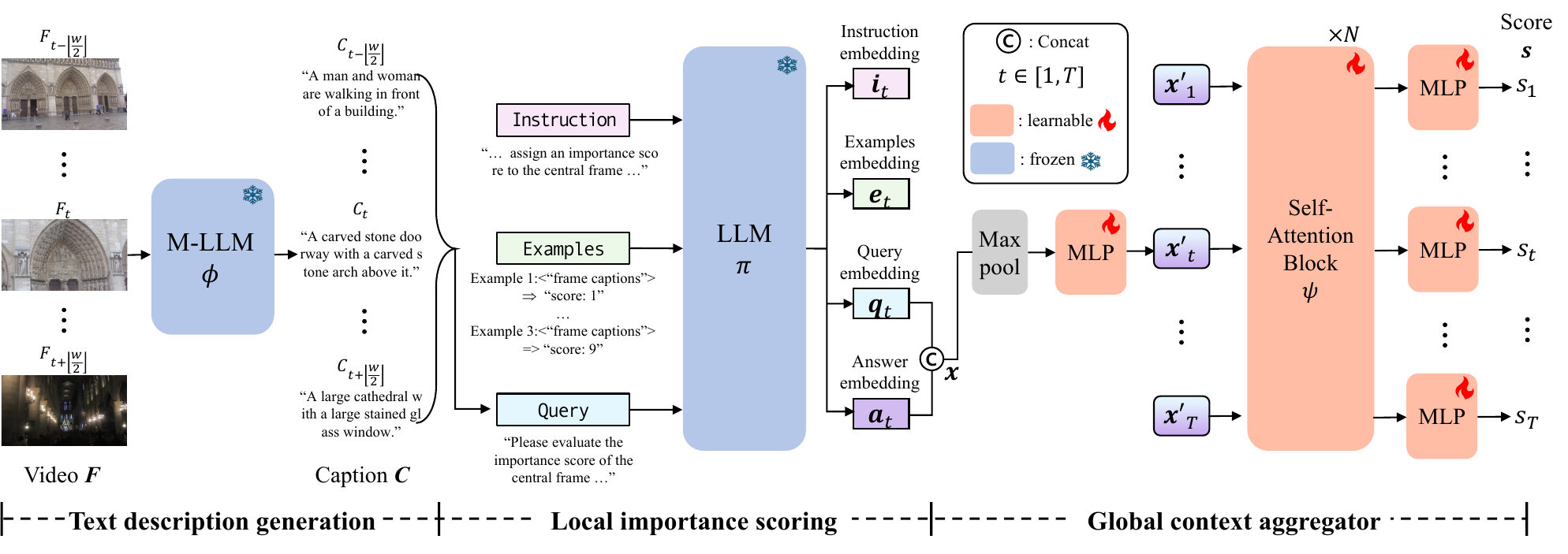}
    \vspace{-0.5cm} 
    \caption{\textbf{Overall architecture.} 
    Our \ours~ framework consists of three key components: text description generation, local importance scoring, and global context aggregation. First, captions for each video frame are generated using a pre-trained Multi-modal Large Language Model (M-LLM)\cite{liu2024visual}. These captions are then incorporated into the query component of an LLM~\cite{touvron2023llama2} by segmenting through a sliding window local context, while instructions and examples are provided as part of the in-context learning prompt.
    We obtain the output embeddings from an intermediate layer of the LLM~\cite{touvron2023llama2}, categorized into instructions, examples, queries, and answers. The query and answer embeddings are pooled and passed through an MLP to produce inputs for the global context aggregator, which encodes the overall context of the input video. Finally, we obtain the output score vectors for the corresponding input video frames.
}
    \vspace{-0.2cm} 

    \label{fig:overall}
\end{figure*}

%% file: sec/3_method.tex
\section{LLM-based Video Summarization (\ours)}
In this section, we present \ours, an LLM-based video summarization framework. Figure~\ref{fig:overall} shows the overall architecture of \ours. \ours~consists of three parts: text description generation via M-LLM, local important frame scoring via LLM, and global context aggregation using self-attention blocks for final predictions of video summarization.

We begin by introducing the problem setup for video summarization in Section~\ref{sec:problem}. Text description generation via M-LLM is discussed in Section~\ref{sec:imagecaption}. Local importance scoring, including in-context learning and extracting embeddings from LLM, is presented in Section~\ref{local_window}. The global context aggregation using self-attention is detailed in Section~\ref{sec:summarizer}, and the training objective is provided in Section~~\ref{sec:training}.

\subsection{Problem Setup}
\label{sec:problem}
Given a video $\bm{F}=[F_1, F_2, \ldots, F_T] \in \mathbb{R}^{T \times H \times W \times 3}$, where $T$ represents the temporal length of the video and $H$ and $W$ denote the height and width of each frame, respectively, the goal of video summarization is to obtain a sequence of importance scores $\bm{s}=[s_1, s_2, \ldots, s_T]\in\mathbb{R}^{T\times 1}$, where higher scores indicate more significant frames.

\subsection{Text Description Generation via M-LLM}
\label{sec:imagecaption}
To incorporate textual data into video summarization, we first generate descriptions of input video frames using a pre-trained M-LLM, denoted by $\phi$~\cite{liu2024visual}.
Specifically, we prompt $\phi$ with ``Provide a detailed one-sentence description,'' generating a sequence of captions $\bm{C} = [C_1, C_2, \ldots, C_T]$,
\begin{align}
    \bm{C} = \phi(\bm{F})
\end{align}
where $C_i$ is the caption for the $i$-th frame.

\subsection{Local Importance Scoring via LLM}\label{local_window}
Given a sequence of captions $\bm{C}$, we employ a pre-trained LLM $\pi$~\cite{touvron2023llama2} to evaluate the importance of each frame within its local temporal context. 
Due to the inherent redundancy in video frames, it is essential to identify key frames based on local context rather than individual frames to filter out repetitive information. To achieve this, we apply a sliding window-based scoring method.
Specifically, for each frame $F_t$ at time-step $t$, the captions within a window of size $w$, denoted as $C_{t-\lfloor\frac{w}{2}\rfloor : t+\lfloor\frac{w}{2}\rfloor}$, are fed into the LLM $\pi$
to evaluate the importance of the central frame $C_t$ in relation to its surrounding frames.
Here, $\lfloor . \rfloor$ denotes the floor function. 
\input{figures/fig_icl}

\paragraph{In-context learning for video summarization.}
To guide the LLM in generating task-specific outputs for video summarization, we apply in-context learning~\cite{zhao2023antgpt, song2023moviechat}, providing examples and instructions directly within the prompt, as shown in Figure~\ref{fig:ICL_prompt}. 
The prompt consists of three components: instructions, examples, and queries.
The instructions define the frame scoring task and criteria; the examples provide three sample question-answer pairs related to video summarization; and the queries contain actual questions for the LLM to answer. The instructions and examples remain fixed, while only the queries vary based on the input. The full prompt configuration is provided in Section~\ref{sec:sup_prompt_llm} at supplementary material. 
In this way, we can effectively leverage the pre-trained LLM for video summarization without finetuning, enabling it to generate consistent and task-specific outputs based on the provided examples and instructions. 

\paragraph{Output embeddings from LLM.}
Rather than obtaining direct answers from the LLM, we extract and utilize output embeddings in video summarization, which retain richer contextual and semantic information from the internal representations. This method offers a potentially more robust assessment of frame importance, preserving essential details that could be abstracted away in final answer outputs.
Specifically, these embeddings are extracted from the Llama-2~\cite{touvron2023llama2} after the RMS Norm layer, as illustrated in Figure~\ref{fig:llama2}.

For each frame $t$, the LLM processes an in-context learning prompt consisting of instruction $\bm{i}$, examples $\bm{e}$, query $\bm{q}$, and answer $\bm{a}$. Since the instructions embeddings $\bm{i}$ and example embeddings $\bm{e}$ remain constant across frames, we focus on the window-specific query embeddings $\bm{q}$ and corresponding answer embedding $\bm{a}$.
As a result, the query embeddings $\bm{q}_t \in \mathbb{R}^{L^{\mathrm{q}} \times D}$ and the answer embeddings $\bm{a}_t \in \mathbb{R}^{L^{\mathrm{a}} \times D}$ are obtained from the LLM $\pi$:
\begin{equation}
\bm{q}_t, \bm{a}_t = \pi(\bm{C}_{t-\lfloor\frac{w}{2}\rfloor : t+\lfloor\frac{w}{2}\rfloor}),
\end{equation}
where $L^{\mathrm{q}}$ and $L^{\mathrm{a}}$ denote the sequence lengths of each embedding, respectively, and $D$ represents the hidden dimension.
Here, $\bm{q}_t$ and $\bm{a}_t$ encode the semantic information relevant to the frame at time-step $t$ within its local context $w$.

\input{figures/fig_llama}
\subsection{Global Context Aggregating via Self-Attention}\label{sec:summarizer}

While the LLM effectively identifies important frames based on local context, incorporating global context is essential for producing a coherent summary of the entire video. To address this, we apply self-attention blocks  $\psi$~\cite{vaswani2017attention}, enabling the model to capture dependencies across the entire video for the final important score prediction.

Within each local window centered at timestep $t$, we first concatenate $\bm{q}_t$ and $\bm{a}_t$ along the sequence axis, producing $\bm{x}_t\in \mathbb{R}^{(L^{\mathrm{q}} + L^{\mathrm{a}}) \times D}$:
\begin{align}
    \bm{x}_t= {\texttt{concat}(\bm{q}_t}, \bm{a}_t).
\end{align}
Then, max pooling is applied to $\bm{x}_t$ along the sequence axis, followed by an $\texttt{MLP}$, resulting in input embeddings $\bm{x}_t' \in \mathbb{R}^{1 \times M}$:
\begin{align}
    \bm{x}_t' = \texttt{MLP}(\texttt{maxpool}(\bm{x}_t)).
\end{align}
The input embeddings for all timesteps, $\bm{x}'=[\bm{x}'_1, ..., \bm{x}'_T]\in\mathbb{R}^{T \times M}$, are provided to the global attention blocks $\psi$:
\begin{align}
\bm{s} &= \texttt{MLP}(\psi (\bm{x}')),
\end{align}
producing the final output of importance scores for the entire video, $\bm{s}\in\mathbb{R}^{T\times1}$.

\subsection{Training Objective}\label{sec:training}
The proposed method is trained using the Mean Squared Error (MSE) loss to optimize frame importance predictions. The loss $\mathcal{L}$ between the ground truth score vector $\hat{\bm{s}}$  and the predicted score $\bm{s}$ and is defined as:
\begin{align}
    \mathcal{L} = \frac{1}{T} \sum_{t=1}^{T} (\textbf{\textit{s}}_t - \hat{\textbf{\textit{s}}}_t)^2.
\end{align}

%% file: figures/fig_icl.tex
\begin{figure}[t]
    \centering
    \includegraphics[width=\linewidth] {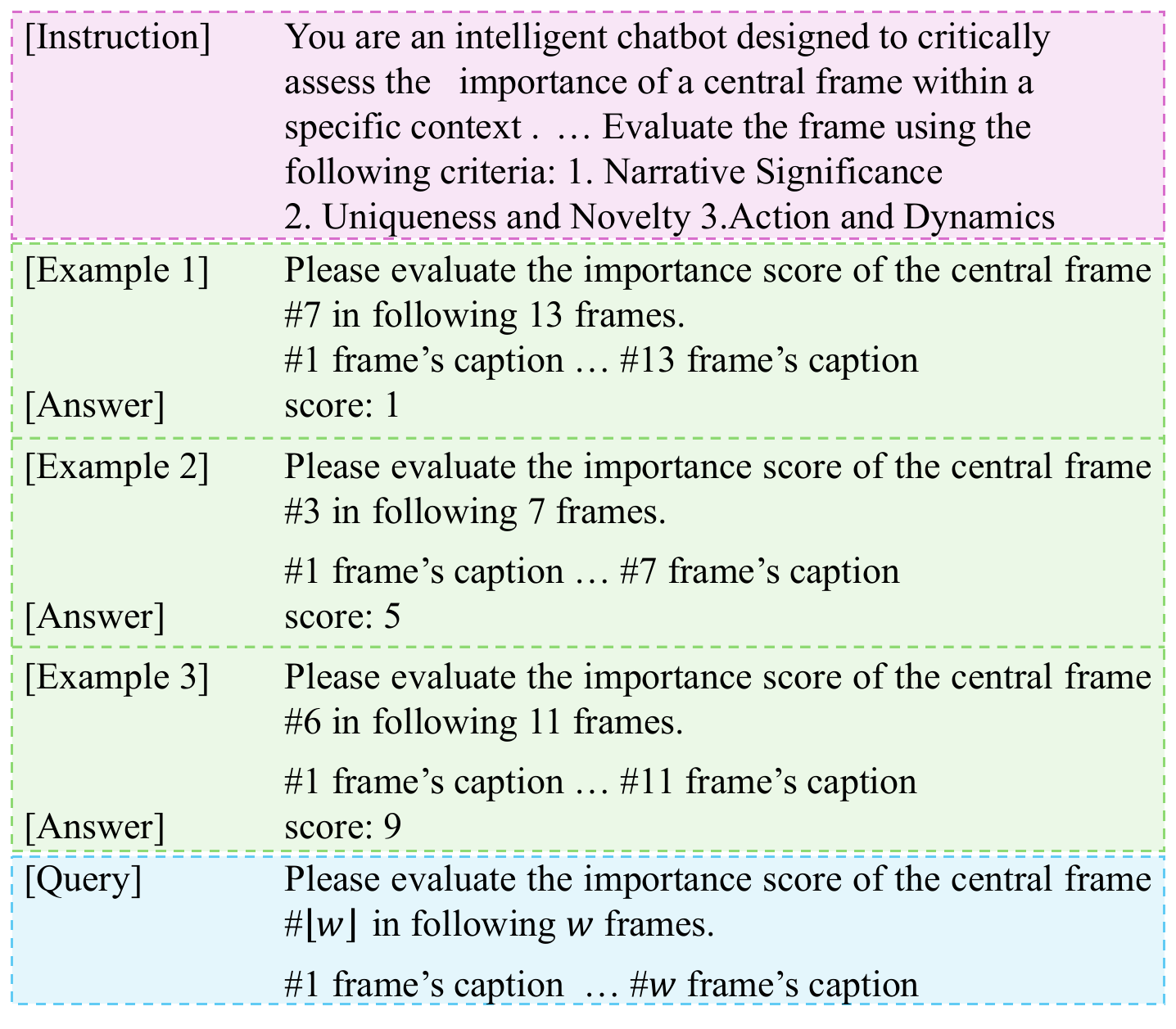}
    \vspace{-0.6cm} 
    \caption{\textbf{In-context learning prompt of LLM.} The instruction for the LLM outlines the video summarization task and specifies the criteria. Then, three examples are provided. Each example includes the number of frame captions and identifies the central frame number as the target. In the query part, the frame captions of our focused video are passed.
    }
    \vspace{-0.4cm}
    \label{fig:ICL_prompt}
\end{figure}

%% file: figures/fig_llama.tex
\begin{figure}[t]
    \centering\scalebox{0.65}{
    \includegraphics[width=\linewidth]{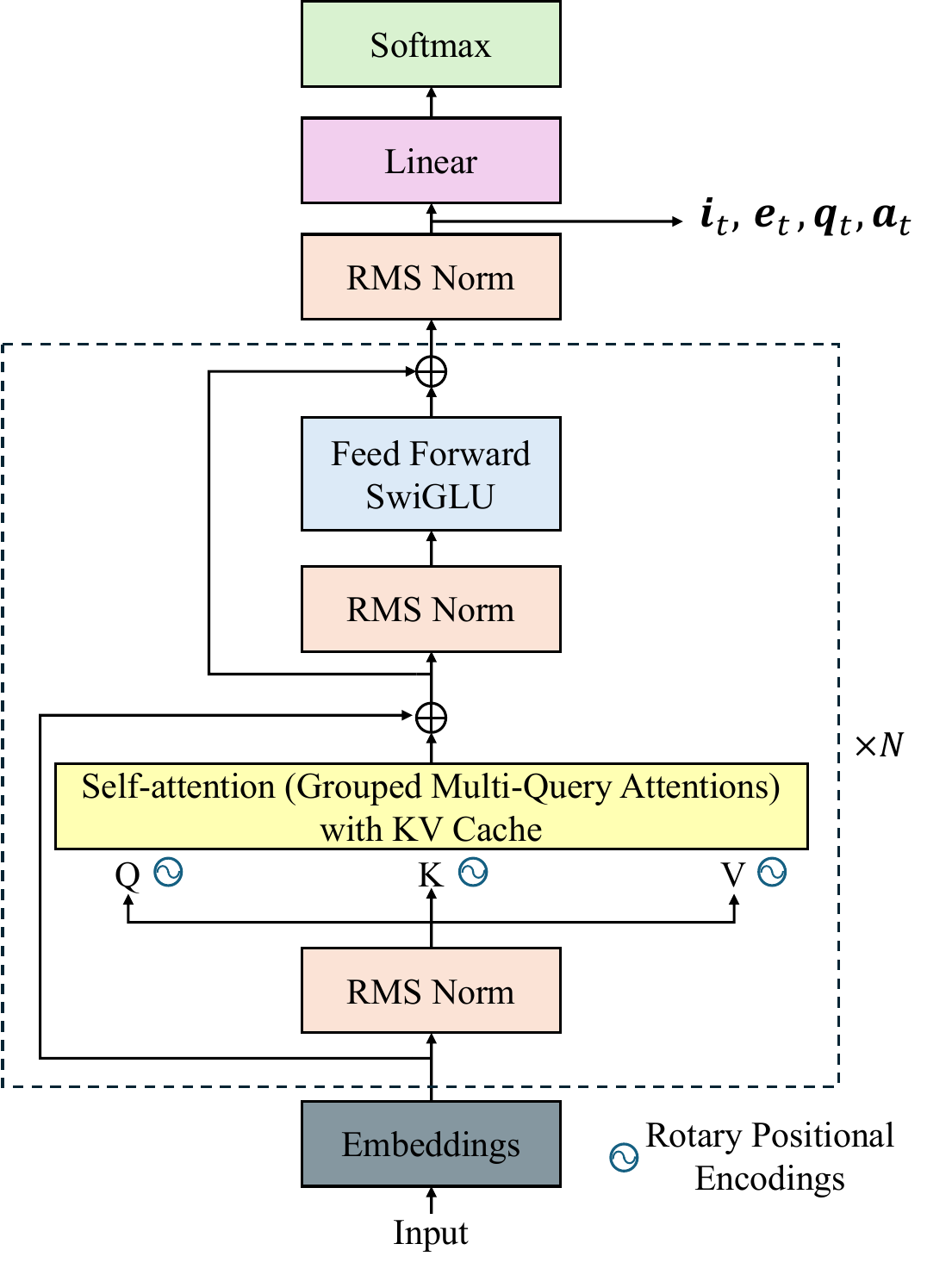}}
    \vspace{-0.1cm} 
    \caption{\textbf{Output embedding from LLM (Llama-2).} Among the output embeddings of instruction $\bm{i}_t$, examples $\bm{e}_t$, query $\bm{q}_t$, and answer $\bm{a}_t$ after the RMS Norm layer of the LLM (Llama-2), we utilize $\bm{q}$ and $\bm{a}$, which retains richer contextual and semantic information of the frame within a local context.
    }
    \label{fig:llama2}
    \vspace{-0.3cm}
\end{figure}

%% file: sec/4_experiments.tex
\section{Experiment} \label{sec:experiment}

\subsection{Datasets}
To evaluate the performance of our method, we use two well-known benchmarks: SumMe~\cite{gygli2014creating} and TVSum~\cite{song2015tvsum}.\\
\textbf{SumMe.} The SumMe~\cite{gygli2014creating} comprises 25 videos, each ranging from 1 to 6 minutes in length, with an average duration of 2 minutes and 40 seconds.  These videos, captured with egocentric, moving, and static cameras, cover various topics, including holidays, events, and sports. Each video is annotated by 15 to 18 raters.\\
\textbf{TVSum.}
The TVSum~\cite{song2015tvsum} includes 50 videos with durations between 2 and 10 minutes, averaging 4 minutes and 11 seconds. This dataset spans diverse content types, such as how-to videos, documentaries, and vlogs. Each video is segmented into equal-length shots, and importance scores are assigned by 20 raters to these segments. 

\subsection{Evaluation Setup} \label{sec:evaluation_setup}

The evaluation protocols for SumMe and TVSum differ in how ground truth and prediction are constructed. In SumMe, the ground truth summary is generated by averaging binary annotations from multiple users. Following the procedure in~\cite{zhang2016video}, we convert the predicted frame-level importance scores  $s$ into video summaries, by aggregating frame scores at the shot level using Kernel Temporal Segmentation (KTS)~\cite{potapov2014category}, which identifies shot boundaries.
Shots with the highest importance scores are selected to form the summary, addressing the 0/1 knapsack problem and ensuring the summary length does not exceed 15\% of the original total duration of the video.
The resulting summary is evaluated against the ground truth summary to measure performance.

For TVSum, importance scores of each user on a continuous 0–1 scale serve as the ground truth.
Evaluation is conducted by comparing the predicted scores $s$ individually with annotations of each user, and the final performance is computed by averaging the results across users.

We evaluate our method using the standard 5-fold cross-validation protocol, following the previous approaches~\cite{apostolidis2021combining, zhu2020dsnet, jiang2022joint, he2023align, son2024csta}.
As evaluation metrics, we adopt rank order statistics, specifically Kendall's $\tau$ and Spearman's $\rho$ following~\cite{otani2019rethinking}.
While the F1 score is widely used in video summarization, it favors short shots over key shots due to length constraints~\cite{otani2019rethinking, terbouche2023multi, son2024csta}.
This limitation may result in an inaccurate reflection of summarization quality. We thus exclude it from our evaluation metrics.

\subsection{Implementation Details}
We employ LLaVA-1.5-7B~\cite{liu2024visual} as the text description generator model $\phi$, and Llama-2-13B-chat~\cite{touvron2023llama2} as the local importance scoring model $\pi$.
The length of each frame caption generated by $\phi$ is limited to a maximum of 77 tokens.
We train our model using the AdamW optimizer~\cite{loshchilov2017decoupled} across $200$ epochs on $5$ NVIDIA A100 GPUs with a batch size of $1$. 
The total training time is approximately 10 hours. 
The learning rate is set to $1.19e-4$ for the SumMe~\cite{gygli2014creating} and $7e-5$ for the TVSum~\cite{song2015tvsum}. 
For both datasets, the window size $w$ is set to $7$ and the dimension reduction $M$ is $2048$. The number of self-attention blocks and the number of heads for global context aggregator $\psi$ are set to 3 and 2, respectively.
The frame captions used in the example section for in-context learning are randomly sampled from the training set of SumMe.

\input{tables/CVPR25_no_f1_tab_sota}

\subsection{Comparison with the State of the Art}
Table~\ref{tab:summe_tvsum} compares the performance of LLM and \ours~with the state-of-the-art models on two benchmark datasets. 
The table is divided into three compartments: (1) random and human baselines, (2) models utilizing use \textit{visual} features, and (3) models using both \textit{visual} and \textit{text} features.
Random and human performance metrics are from~\cite{otani2019rethinking}, where random performance is computed by averaging the results of comparisons between 100 randomly generated value sequences in the range $[0,1]$ and the ground truth. 

\noindent
\textbf{LLMs.}
Table~\ref{tab:summe_tvsum} investigates the impact of the general knowledge of LLM~\cite{touvron2023llama2} on video summarization by evaluating it in a zero-shot setting. Importance scores are obtained via in-context learning (Figure~\ref{fig:ICL_prompt}), using the same experimental setup as \ours, but without the global context aggregator $\psi$. 
The LLM achieves competitive performance among previous methods on SumMe, demonstrating the effectiveness of leveraging textual data alongside its general knowledge. In contrast, its performance on TVSum is comparatively lower. This discrepancy can be attributed to differences in the evaluation protocols of SumMe and TVSum, as described in Section~\ref{sec:evaluation_setup}. In SumMe, evaluation is performed by averaging the user summaries, whereas in TVSum, evaluation is conducted separately for each user score, and the results are then averaged. 
This performance gap, attributed to differences in subjectivity between the two datasets, indicates that the LLM is well-suited for general summarization tasks but less effective in capturing individual user preferences.

\noindent
\textbf{\ours.}
Our model achieves state-of-the-art performance on both benchmark datasets.
Notably, \ours~shows significant performance gains over the zero-shot LLM, as seen by comparing the last two rows in Table~\ref{tab:summe_tvsum}. The result indicates that the proposed method effectively handles both general and subjective aspects of keyframe selection in video summarization.
In particular, it highlights the importance of the global context aggregator $\psi$, which enhances the reasoning ability of LLM by capturing contextual relationships across local windows, enabling more coherent and informed sequence-level decision-making.
Moreover, our method outperforms existing multimodal video summarization models~\cite{he2023align, narasimhan2021clip, argaw2024scaling, li2023progressive}, where text information serves as an auxiliary input to support visual processing. 
Unlike these approaches, our framework centers summarization around language, relying on textual descriptions and the reasoning capabilities of LLMs. These results underscore the advantage of integrating textual data with the broad reasoning capabilities of LLMs, enabling more contextually aware and semantically rich video summaries.

\subsection{Analysis}

\input{tables/CVPR25_no_f1_tab_finetuning}

\paragraph{Finetuning (M-)LLM, $\phi$ and $\pi$.}
To determine whether the performance improvements arise from fundamental architectural enhancements or are simply due to finetuning on downstream benchmarks, we conduct experiments on both zero-shot and finetuned settings using (M-)LLM.

We first establish a baseline by evaluating M-LLM (LLaVA~\cite{liu2024visual}) and LLM (Llama-2~\cite{touvron2023llama2}) in zero-shot settings.
In particular, we assess whether M-LLM can directly serve as the importance scorer $\pi$, assigning frame-level importance scores without relying on captions, as shown in the first row of Table~\ref{tab:reb_mllm_finetuning}. 
Comparing the first and third rows of the table demonstrates that explicitly providing captions from M-LLM to the LLM yields better results than direct scoring by M-LLM alone, underscoring the importance of leveraging language semantics in video summarization.

Subsequently, to evaluate the impact of finetuning, we apply LoRA~\cite{zheng2024llamafactory} to both the M-LLM and LLM, with the finetuned models denoted by $^{*}$ in Table~\ref{tab:reb_mllm_finetuning}. 
The performance improvements observed when comparing the first and second rows, as well as the third and fourth rows, validate the effectiveness of finetuning. 
However, LLMVS exhibits significantly greater improvements in the fifth row compared to these baselines, demonstrating that its effectiveness extends beyond simple finetuning.

\input{tables/CVPR25_no_f1_tab_prompting}

\paragraph{Prompting to (M-)LLM , $\phi$ and $\pi$.}
Prompting is essential in (M-)LLMs, as it determines how the model processes information and generates responses. To evaluate the effectiveness of different prompting strategies, we examine prompts for both M-LLM $\phi$ and LLM $\pi$.

For M-LLM $\phi$, we explore the impact of different captioning styles. Specifically, we examine how the richness and descriptiveness of frame captions influence summarization. As detailed in Section~\ref{sec:imagecaption},  we instruct the M-LLM with a generic prompt ``Provide a detailed one-sentence description." To obtain more fine-grained descriptions, we instruct the model to separately describe the center and background regions of the image using two prompts: ``Describe the center part of this image in one detailed sentence" and ``Describe the background part of this image in one detailed sentence."
Table~\ref{tab:prompting_comparison} (a), the generic prompt yields better results than the central-background approach. This suggests that a broader, high-level description allows the model to better capture scene dynamics and key events, reducing reliance on specific frame regions.

For LLM $\pi$, we compare two prompting types: (1) direct numerical scoring of frame importance using the prompt, ``Please evaluate the importance score of the central frame in following frames," as described in Figure~\ref{fig:ICL_prompt}; and (2) textual explanation, where the LLM is instructed to summarize frame captions within a local window using the prompt, ``Please summarize the following frames in one sentence." inspired by~\cite{islam2024video}. Table~\ref{tab:prompting_comparison} (b) shows that direct numerical scoring consistently outperforms textual summarization, suggesting that assigning explicit importance scores provides a clearer and more effective evaluation of frame significance.

\input{tables/CVPR25_no_f1_tab_ablations}
\paragraph{Ablation studies.}
We conduct experiments to validate the effectiveness of using the output embeddings $\bm{q}$ and $\bm{a}$ and self-attention blocks $\psi$. Table~\ref{tab:query_score} shows the results. 
From the first and second rows, we either use query embeddings $\bm{q}$ or answer embeddings $\bm{a}$ during input structuring in Section~\ref{sec:summarizer}.
Comparing the first and second rows shows that leveraging query embeddings yields better performance than leveraging answer embeddings alone, highlighting the importance of contextual information in assessing frame relevance and enriching the semantic processing capabilities of the LLM. Furthermore, the third row, which combines both query and answer embeddings with the global context aggregator, achieves the best results, confirming that integrating both query and answer embeddings with the global context aggregator yields the most effective results.
In the fourth row, we replace the self-attention blocks (SAB) used as $\psi$ with a simple MLP.
A comparison between the third and fourth rows demonstrates that employing the global self-attention block is more effective than using an MLP.

\input{figures/fig_qual}
\input{tables/CVPR25_no_f1_tab_llama_emb}

\paragraph{Extraction position of output embeddings $\bm{q}$ and $\bm{a}$.}
In Table~\ref{tab:llm_emb}, we examine the effects of extraction position of output embeddings $\bm{q}$ and $\bm{a}$. Specifically, we consider two positions within the LLM $\pi$: after the RMS Norm layer and after the Linear layer, as shown in Figure~\ref{fig:llama2}. Since embeddings extracted after the linear layer are specialized for word domains, we aim to explore the effectiveness of embeddings obtained both before and after this specialization, namely after the RMS Norm layer and the Linear layer, respectively.

Table~\ref{tab:llm_emb} presents that embeddings extracted after the RMS Norm layer outperform those after the Linear layer, likely due to their retention of richer contextual information, whereas embeddings after the Linear layer are more specialized for word domains.

\input{tables/CVPR25_no_f1_tab_zeroshot}
\paragraph{Zero-shot evaluation on MR.HiSum.}
To evaluate the generalization ability of the proposed method on unseen videos, we train \ours~on the SumMe~\cite{gygli2014creating} and test its zero-shot performance on a random subset of 50 videos from the MR.HiSum~\cite{sul2024mr}. 
We compare \ours~with previous methods~\cite{apostolidis2021combining, vaswani2017attention, zhu2020dsnet}.
Table~\ref{tab:zeroshot} shows that \ours~outperforms other models, demonstrating its strong generalization capability in zero-shot settings.
This result suggests that by leveraging the advanced capability of the LLM to interpret text-based information and incorporating its contextual embeddings, \ours~effectively captures representations that extend well to unseen video content.

\subsection{Qualitative Results}
Figure~\ref{fig:qual} presents qualitative results on TVSum~\cite{song2015tvsum}, comparing predicted scores from \ours~against the ground-truth user scores. 
The x- and y-axes are time step $t$ and importance score $s$, respectively. In this figure, the blue line represents the average user scores, while the orange line shows the normalized predicted scores from our model. All scores are in the range of 0 to 1. 
Green areas indicate segments that received high importance scores, while pink areas correspond to segments with low scores.

The predicted importance scores align well with overall trends in the ground truth, highlighting the robustness and generalization capability of our approach. We also observe that action-related scenes tend to receive higher importance scores from both human annotators and \ours.
For example, in Figure~\ref{fig:qual} (a), which presents a video about instructing how to stop a bike, scenes where a woman is talking to the camera receive relatively low scores. In contrast, frames depicting dynamic actions—such as riding or touching the bike—are assigned higher scores. Similarly, in Figure~\ref{fig:qual} (b), which features a motorcycle stunt show, frames showing a man being interviewed are rated lower, whereas those involving high-energy activities, such as stunts, receive higher scores. These patterns suggest that \ours~effectively identifies and emphasizes action-oriented content that contributes significantly to the narrative.

%% file: tables/CVPR25_no_f1_tab_sota.tex
\begin{table}[t!]
\centering
\begin{tabular}{@{}lC{1.5cm}C{1cm}C{1cm}C{1cm}@{}}
\toprule
\multirow{2}{*}{Method} & \multicolumn{2}{c}{SumMe} & \multicolumn{2}{c}{TVSum~} \\ \cmidrule(l){2-3} \cmidrule(l){4-5}
                & \(\tau\) & \(\rho\)  & \(\tau\) & \(\rho\) \\ \midrule
Random~\cite{otani2019rethinking}  & 0.000 & 0.000  & 0.000 & 0.000 \\
Human~\cite{otani2019rethinking}  & 0.205 & 0.213  & 0.177 & 0.204 \\ \midrule
\textit{Visual} \\
VASNet~\cite{fajtl2019summarizing} & 0.160 &  0.170 & 0.160 & 0.170 \\
DSNet-AB~\cite{zhu2020dsnet}  & 0.051 &  0.059 & 0.108 &  0.129 \\
DSNet-AF~\cite{zhu2020dsnet}  & 0.037 &  0.046  & 0.113 &  0.138 \\
DMASum~\cite{wang2020query} &0.063 &0.089 &0.203 &0.267 \\
PGL-SUM~\cite{apostolidis2021combining} &- &- &0.206	&0.157\\
MSVA~\cite{ghauri2021supervised} &0.200 &0.230 &0.190 &0.210 \\
iPTNet~\cite{jiang2022joint}  & 0.101 &  0.119  & 0.134 &  0.163\\ 
CSTA~\cite{son2024csta}  & \underline{0.246} &  \underline{0.274}  & \underline{0.194} &  \underline{0.255} \\ \midrule
\textit{Visual + Text} \\
CLIP-It~\cite{narasimhan2021clip} &- &- &0.108 &0.147 \\
A2Summ~\cite{he2023align}  & 0.108 &  0.129  & 0.137 &  0.165 \\
SSPVS~\cite{li2023progressive} &0.192 &0.257 &0.181 &0.238 \\
Argaw \etal~\cite{argaw2024scaling}  & 0.130  & 0.152  & 0.155  &  0.186 \\
\rcol LLM & 0.170 & 0.189   &0.051  & 0.056 \\ 
\rcol \textbf{\ours~(ours)}  & \textbf{0.253} & \textbf{0.282} & \textbf{0.211} & \textbf{0.275} \\
\bottomrule
\end{tabular}%
\vspace{-0.1cm}
\caption{\textbf{Comparison with the state of the arts.} 
The table is divided into three compartments: (1) random and human baselines, (2) models utilizing use visual features, and (3) models using both visual and text features. \ours~achieves the state-of-the-art performance on two benchmark datasets.
}
\vspace{-0.5cm}
\label{tab:summe_tvsum}
\end{table}	

%% file: tables/CVPR25_no_f1_tab_finetuning.tex
\begin{table}[t]
    \centering
    \scalebox{1.0}{
    \begin{tabular}{L{0.5cm}C{1.3cm}C{1.3cm}C{0.9cm}C{0.9cm}C{0.9cm}}
    \toprule
    &$\phi$ & $\pi$ & $\psi$ & \(\tau\) & \(\rho\) \\ \midrule
    (1) &-  & LLaVA & -  & 0.119  & 0.132 \\
    (2) &- & LLaVA$^*$ & -  & 0.140 & 0.156\\
    (3) &LLaVA & Llama & - & 0.170 & 0.189 \\
    (4) &LLaVA & Llama$^*$ &-   & 0.181 & 0.201 \\
    \rcol (5) &\textbf{LLaVA} & \textbf{Llama} & \textbf{SAB$^*$} & \textbf{0.253} & \textbf{0.282} \\
    \bottomrule
    \end{tabular}
    }
    \vspace{-0.1cm}
    \caption{\textbf{Finetuning (M-)LLM, $\phi$ and $\pi$.} $\phi$: text generator, $\pi$: local importance scorer, $\psi$: global context aggregator, *:finetuned, SAB: self-attention blocks.}
    \label{tab:reb_mllm_finetuning}
\end{table}

%% file: tables/CVPR25_no_f1_tab_prompting.tex
\begin{table}[t]
\centering
\begin{subtable}[t]{0.45\textwidth}
\centering
\begin{tabular}{@{}L{4cm}C{1.3cm}C{1.3cm}@{}}
\toprule
Prompt type & \(\tau\) & \(\rho\) \\ \midrule
Central-background & 0.241 & 0.269 \\
\rcol \textbf{Generic} & \textbf{0.253} & \textbf{0.282} \\
\bottomrule
\end{tabular}
\caption{Prompting to M-LLM $\phi$}
\end{subtable}
\begin{subtable}[t]{0.45\textwidth}
\centering
\begin{tabular}{@{}L{4cm}C{1.3cm}C{1.3cm}@{}}
\toprule
Prompt type & \(\tau\) & \(\rho\) \\ \midrule
Textual explanation & 0.239 & 0.266 \\
\rcol \textbf{Numerical evaluation} & \textbf{0.253} & \textbf{0.282} \\
\bottomrule
\end{tabular}
\caption{Prompting to LLM $\pi$}
\end{subtable}
\vspace{-0.1cm}
\caption{\textbf{Prompting to (M-)LLM.} Evaluation of different prompting styles applied to (a) M-LLM $\phi$ and (b) LLM $\pi$ on the SumMe dataset~\cite{gygli2014creating}. All experiments use a window size of \(w = 7\).}
\label{tab:prompting_comparison}
\vspace{-0.3cm}
\end{table}

%% file: tables/CVPR25_no_f1_tab_ablations.tex
\begin{table}[t]
\centering

\scalebox{1.0}{
\begin{tabular}{@{}lccccc@{}}
\toprule
&Query $\bm{q}$& Answer $\bm{a}$& $\psi$ & \(\tau\) & \(\rho\) \\ \midrule
(1)&-&\checkmark & SAB$^*$ & 0.233 & 0.260 \\
(2)&\checkmark&- &SAB$^*$ & 0.238 & 0.265 \\
\rcol (3)&\textbf{\checkmark}&\textbf{\checkmark} & \textbf{SAB$^*$} & \textbf{0.253} &  \textbf{0.282}  \\
(4)&\checkmark&\checkmark & MLP$^*$ &0.182  &0.203 \\
\bottomrule
\end{tabular}
}
\vspace{-0.1cm}
\caption{\textbf{Ablation studies.} The embeddings are used individually or concatenated. Performance is evaluated on the SumMe dataset~\cite{gygli2014creating}.
*:finetuned, MLP: MLP only (without self-attention blocks), SAB: self-attention blocks.
}
\vspace{-0.4cm} 
\label{tab:query_score}
\end{table}

%% file: figures/fig_qual.tex
\begin{figure*}[ht!]
    \centering
    \scalebox{0.95}{
    \includegraphics[width=\linewidth]{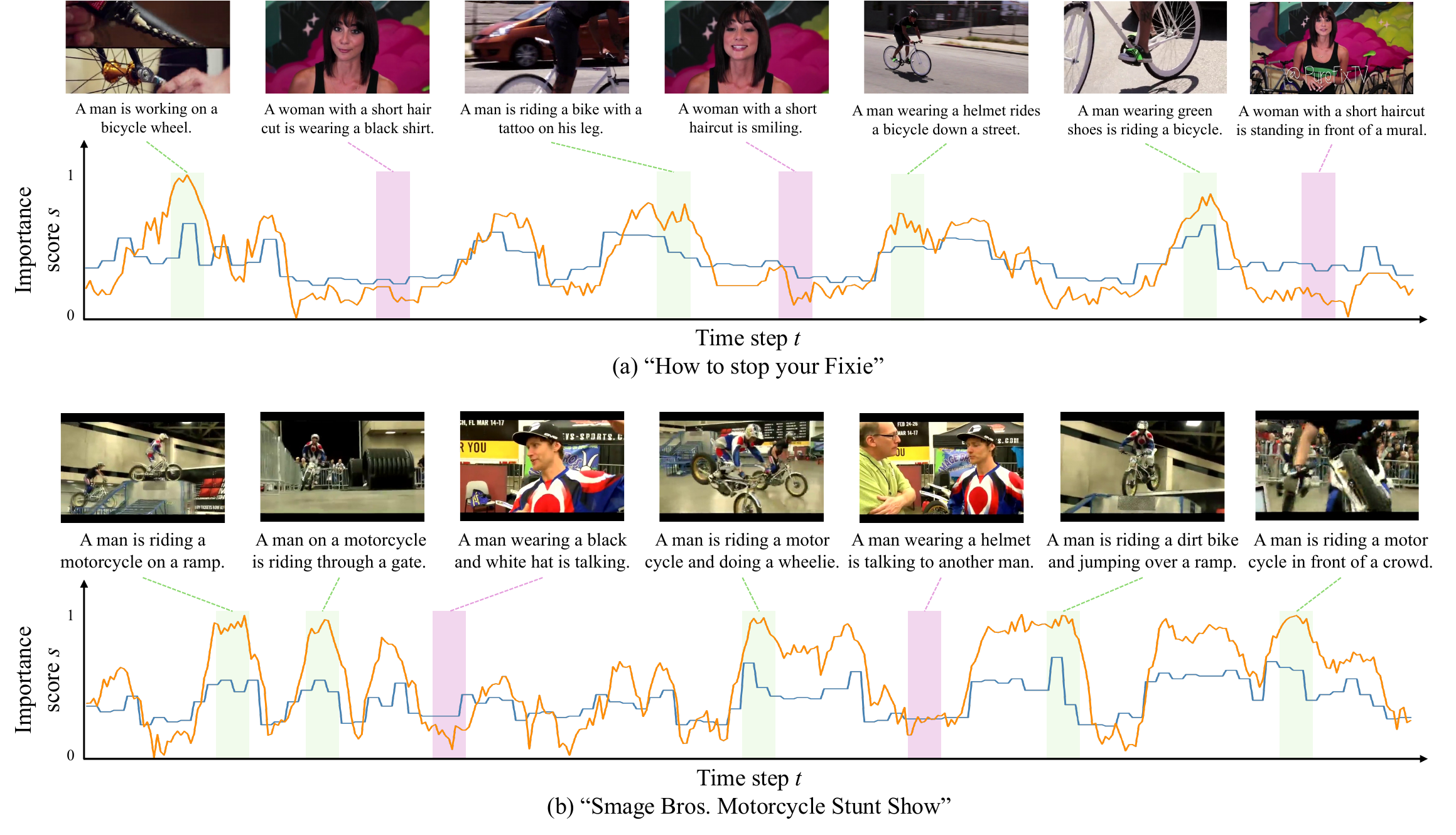}
    }
    \vspace{-0.4cm}
    \caption{\textbf{Qualitative results.}
    Videos are from the TVSum dataset~\cite{song2015tvsum}. The x-axis and y-axis represent time step $t$ and importance score $s$, respectively. The blue line indicates the averaged user scores from the ground truth annotations, while the orange line denotes the predicted importance scores from our model, normalized to the range [0, 1]. Green regions highlight segments where importance scores are high, whereas pink regions indicate segments where importance scores are low.}
    \vspace{-0.3cm} 
    
    \label{fig:qual}
\end{figure*}

%% file: tables/CVPR25_no_f1_tab_llama_emb.tex
\begin{table}[t]
\centering

\begin{tabular}{@{}L{4cm}C{1.3cm}C{1.3cm}@{}}
\toprule
Extraction position & \(\tau\) & \(\rho\) \\ \midrule
After Linear layer & 0.241 & 0.269  \\
\rcol \textbf{After RMS Norm layer} & \textbf{0.253} & \textbf{0.282}\\
\bottomrule
\end{tabular}
\vspace{-0.2cm}
\caption{\textbf{Extraction position of output embeddings $\bm{q}$ and $\bm{a}$.} Evaluation performed on SumMe with window size $w = 7$.}
\vspace{-0.2cm} 
\label{tab:llm_emb}
\end{table}

%% file: tables/CVPR25_no_f1_tab_zeroshot.tex
\begin{table}[t]
\centering

\begin{tabular}{@{}L{3.5cm}C{1.5cm}C{1.5cm}@{}}
\toprule
Method & \(\tau\) & \(\rho\)  \\ \midrule
VASNet~\cite{vaswani2017attention} &0.364 &0.364 \\
PGL-SUM~\cite{apostolidis2021combining}  &0.375 &0.375 \\
DSNet-AB~\cite{zhu2020dsnet}  &0.362 &0.362 \\
DSNet-AF~\cite{zhu2020dsnet}  &0.342 &0.342\\ 
\rcol \textbf{\ours~(ours)} & \textbf{0.440} & \textbf{0.440}\\
\bottomrule
\end{tabular}
\vspace{-0.0cm}
\caption{\textbf{Zeroshot evaluation on MR.HiSum.} The evaluation is conducted on 50 randomly selected videos from the MR.HiSum~\cite{sul2024mr}. Both previous methods and LLMVS are tested directly on MR.HiSum using pre-trained models which are originally trained on SumMe.}
\vspace{-0.5cm}
\label{tab:zeroshot}
\end{table}

%% file: sec/5_conclusion.tex
\section{Conclusion} \label{sec:conclusion}
We have introduced the LLM-based Video Summarization framework (\ours), which leverages the semantic understanding capabilities of large language models to perform video summarization through caption-guided frame scoring.
LLMVS integrates textual descriptions generated by M-LLM from video frames, which are then evaluated and refined through the LLM using a comprehensive local-global context aggregation network. This design allows the model to capture narrative structure more effectively by combining the descriptive strength of the M-LLM with the reasoning capabilities of the LLM. Experiments on the SumMe and TVSum demonstrate the effectiveness of the proposed approach, showing consistent improvements over existing methods.
By bridging the gap between visual data and language, LLMVS enhances the summarization process and sets a new direction for future research in multimedia content analysis, enabling advanced cross-modal reasoning.

%% file: sec/6_ack.tex
\smallbreak
\noindent \textbf{Acknowledgements.}
\scriptsize{This work was supported by the NRF grant (RS-2021-NR059830 (35\%)), the IITP grants (RS-2022-II220264: Comprehensive Video Understanding and Generation (40\%), RS-2019-II191906: Artificial Intelligence Graduate School Program at POSTECH (5\%)) funded by the Korea government (MSIT), the DIPS1000+ grant (20266807: Generative AI Core Technology for Creating High-Quality Synthetic Data and Its Application to Vision AI System for Autonomous Driving) by KISED (10\%), and the Defense Innovation Company 100 Exclusive R\&D Support Program (R240202: Development of Sensor-Level High-Quality Multimodal/Video Synthesis Data Generation and Real-time Target Recognition Technology for Unmanned System) by the Korea goverment (KRIT) (10\%).}

%% file: sec/appendix.tex
\clearpage
\maketitlesupplementary
\appendix
\normalsize
\label{sec:appendix}
\vspace{-3mm}
\renewcommand\thesection{\Alph{section}}
\renewcommand\thefigure{\Alph{section}\arabic{figure}}
\renewcommand\thetable{\Alph{section}\arabic{table}}

\setcounter{figure}{0}
\setcounter{table}{0}

In this appendix, we provide additional experiments in Section~\ref{sec:sup_exp}, additional qualitative results in Section~\ref{sec:sup_qual}, and full prompts for LLM~$\pi$ in Section~\ref{sec:sup_prompt_llm}.

\section{Additional Experiments}
\label{sec:sup_exp}
\paragraph{Model choice of (M-)LLM, $\phi$ and $\pi$.}
While LLaVA~\cite{liu2024visual} generates captions based on single-frame inputs, Video-LLaVA~\cite{lin2023video} processes multiple frames simultaneously, enabling video-level understanding. Given the flexibility of our framework in M-LLM selection, we incorporate both models to examine how the design of the captioning model influences downstream performance.

We exploit Video-LLaVA as both the text generator $\phi$ and the local importance scorer $\pi$ in the last row of Table~\ref{tab:reb_videollava}. In this setup, we first obtain captions from Video-LLaVA and then feed both the generated captions and the corresponding video frames back into Video-LLaVA for local window-based scoring, where the model supports a maximum of 8 frames per window.
Table~\ref{tab:reb_videollava} presents a comparison between two configurations, showing similar performance. This suggests that once textual descriptions are well aligned with the video content, additional visual input at the scoring stage offers limited benefit. Moreover, modeling local temporal context via $\pi$ over sequential captions appears sufficient to compensate for the limitations of frame-level captioning.

\paragraph{Effects of the window size $w$.}
To determine the optimal window size $w$, we examine the impact of varying $w$ on summarization performance. An appropriate window size should provide enough contextual information without introducing excess detail, balancing local context for effective summarization. As shown in Table~\ref{tab:window_size}, the LLMVS model achieves its best performance at $w=7$, suggesting that this window size offers the ideal balance between capturing local context and maintaining coherence for summarization.

\paragraph{Effects of the number of self-attention blocks.}
To evaluate the impact of self-attention depth in the global context aggregator, we examine configurations with 2, 3, and 4 self-attention blocks. As shown in Table~\ref{tab:layers}, using three self-attention blocks improves performance compared to two self-attention block, likely due to enhanced contextual integration. However, performance decreases when increasing to four self-attention blocks, possibly due to added complexity. Thus, three self-attention blocks provides the best balance for video summarization.

\input{tables/CVPR25_supp_no_f1_tab_videollava}

\input{tables/CVPR25_supp_no_f1_tab_window}
\input{tables/CVPR25_supp_no_f1_tab_layer}

\section{Additional Qualitative Results}
\setcounter{figure}{0}
\setcounter{table}{0}
\label{sec:sup_qual}
\input{figures/supp_fig_qual}

Figure~\ref{fig:supp_qual} presents additional qualitative results, comparing the summaries generated by \ours~with the ground truth on (a) the SumMe\cite{gygli2014creating} and (b) the TVSum~\cite{song2015tvsum} datasets.
In SumMe, the x-axis denotes the time step $t$, while the y-axis represents the binarized summary. The blue line shows the averaged binary summaries from multiple users, and the orange line represents our predicted summary, obtained by applying the KTS and 0/1 knapsack algorithm to the predicted frame scores. As shown in Figure~\ref{fig:supp_qual}(a), the predicted summaries closely align with the peaks in the ground truth. For example, \ours~successfully identifies key transitions, such as when the camera falls to the ground or when a car drives over a ground-level camera.
Figure~\ref{fig:supp_qual}(b) illustrates the results on TVSum. Here, the x-axis again represents the time step $t$, and the y-axis indicates importance scores. The blue line shows the average of user-provided scores ranging from 0 to 1, while the orange line represents normalized predicted scores of \ours. Both human annotations and our predictions exhibit similar trends—higher scores are assigned to action-oriented segments (e.g., working on or touching a tire), while lower scores are given to static or less informative scenes.
By leveraging the local window of captions, \ours~effectively captures the narrative context of shots and identifies critical contents, aligning closely with human perception of scene importance. These results further demonstrate the robustness and generalization capability of \ours~across diverse user annotations and video summarization benchmarks.

\section{Full Prompts for LLM $\pi$}
\setcounter{figure}{0}
\setcounter{table}{0}
\label{sec:sup_prompt_llm}
In this section, we provide full prompts given to LLM $\pi$ for in-context learning in Table~\ref{tab:full_llm_prompt}.
As illustrated in Figure~\ref{fig:ICL_prompt}, our prompts consist of three parts: instruction $\bm{i}$, examples $\bm{e}$, and queries $\bm{q}$.
The instructions guide LLM regarding the video summarization task, followed by three examples.
Each example includes a question-answer pair, where the question requests score evaluations with frame captions, and the answers, ranging from one to ten, are derived from the dataset.
The queries are direct questions given to LLM, requiring the desired actual answers $\bm{a}$.

\begin{onecolumn}
\input{tables/CVPR25_supp_prompt_llm}
\end{onecolumn}

%% file: tables/CVPR25_supp_no_f1_tab_videollava.tex
\begin{table}[t]
    \centering
    \scalebox{1.0}{
    \begin{tabular}{cccccc}
    \toprule
    $\phi$ & $\pi$ & $\psi$ & \(\tau\) & \(\rho\) \\ \midrule
    LLaVA & Llama & SAB$^*$ & \textbf{0.253} & \textbf{0.282} \\
    Video-LLaVA & Video-LLaVA & SAB$^*$   &0.252	  &0.281  \\
    \bottomrule
    \end{tabular}
    }
    \vspace{-2mm}
    \caption{\textbf{Model choice of (M-)LLM, $\phi$ and $\pi$.} $\phi$: text generator, $\pi$: local importance scorer, $\psi$: global context aggregator, *:finetuned, SAB: self-attention blocks.}
    \vspace{-2mm}
    \label{tab:reb_videollava}
\end{table}

%% file: tables/CVPR25_supp_no_f1_tab_window.tex
\begin{table}[t]
\centering

\begin{tabular}{@{}lC{1.2cm}C{1.2cm}C{1.2cm}@{}}
\toprule
Method & $w$  & \(\tau\) & \(\rho\) \\ \midrule
\ours~(ours) &5  & 0.236 & 0.263  \\
\rcol \ours~(ours) &7   &\textbf{0.253} & \textbf{0.282}  \\
 \ours~(ours) &9  & 0.245  & 0.274  \\
\bottomrule
\end{tabular}
\caption{\textbf{Effects of Window Size $w$.} Evaluation performed on the SumMe dataset~\cite{gygli2014creating} with 3 self-attention blocks and 2 multi-head attention heads.}
\label{tab:window_size}
\end{table}

%% file: tables/CVPR25_supp_no_f1_tab_layer.tex
\begin{table}[t]
\centering

\begin{tabular}{@{}C{1.5cm}C{1.5cm}C{1.5cm}@{}}
\toprule
$N$ & \(\tau\) & \(\rho\) \\ \midrule
2 & 0.243 & 0.271\\
\rcol 3  & \textbf{0.253} & \textbf{0.282}  \\
4  &0.244  &0.272 \\
\bottomrule
\end{tabular}
\caption{\textbf{Effects of the Number of Self-Attention Blocks ($N$).} Evaluation performed on the SumMe dataset~\cite{gygli2014creating} with 2 multi-head attention heads.}
\label{tab:layers}
\end{table}

%% file: figures/supp_fig_qual.tex
\begin{figure*}[ht!]
    \centering
    \scalebox{1.0}{
    \includegraphics[width=\linewidth]{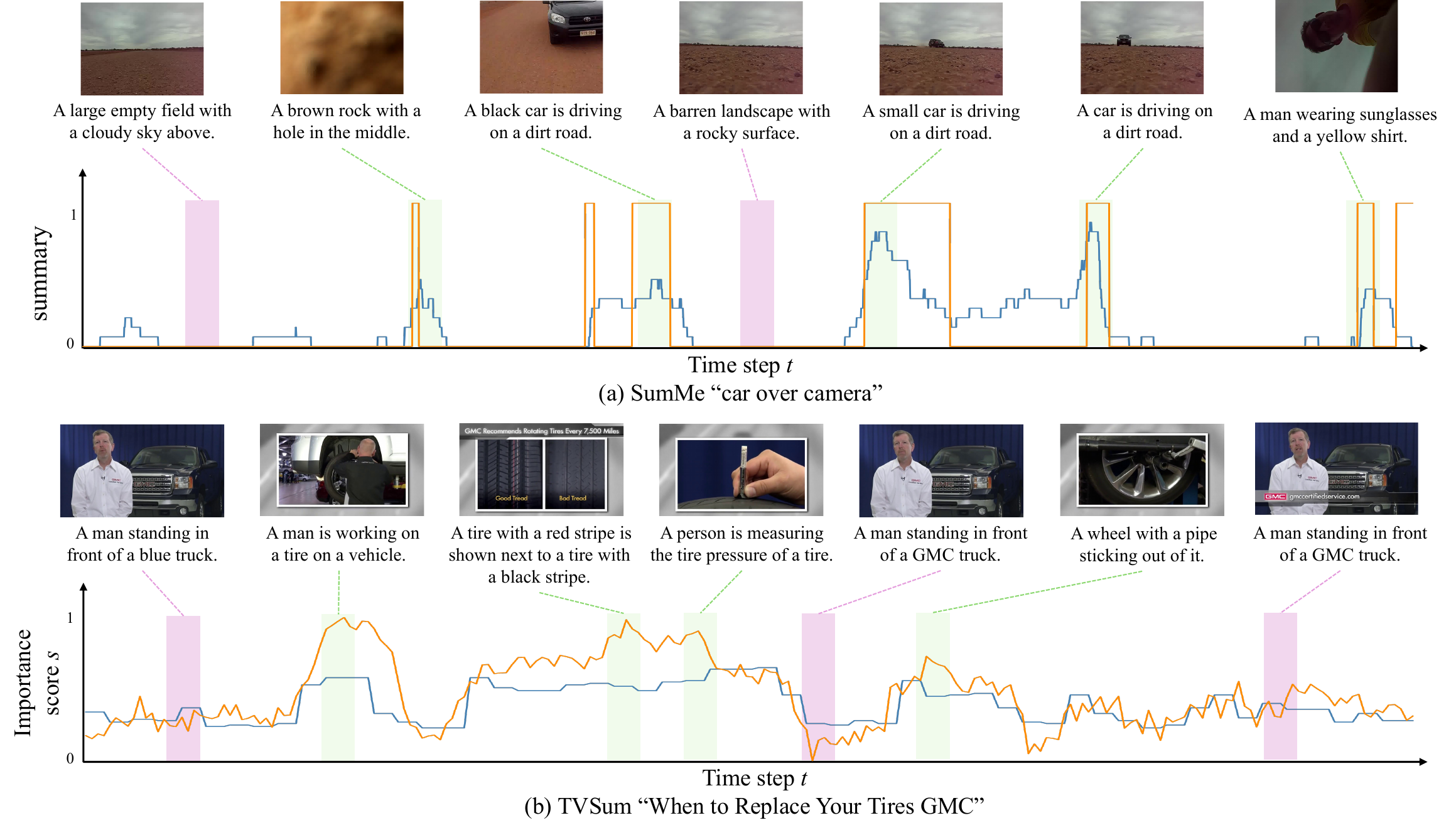}}
    \vspace{-0.7cm}
    \caption{\textbf{Additional qualitative results.} Green regions highlight segments where importance scores are high, whereas pink regions indicate segments where importance scores are low. (a) Results from SumMe~\cite{gygli2014creating}. The x-axis and y-axis represent time step $t$ and binarized summary, respectively. The blue line represents the average of binary user summaries in the ground truth, and the orange line is the predicted summary of our model, which is processed using the KTS and 0/1 knapsack algorithm on predicted frame score. (b) Results from TVSum~\cite{song2015tvsum}. The x-axis and y-axis represent time step $t$ and importance score $s$, respectively. The blue line is the average of user scores ranging in $[0, 1]$, and the orange line is the normalized predicted importance score.}
    \label{fig:supp_qual}
\end{figure*}

%% file: tables/CVPR25_supp_prompt_llm.tex
\begin{longtable}{p{1.5cm}|p{13cm}}
\toprule
Instruction & You are an intelligent chatbot designed to critically assess the importance of a central frame within a specific context. Given a set of consecutive frame descriptions from a video with narrative changes, your task is to assign an importance score to the central frame based on its narrative contribution. Evaluate the frame using the following criteria: \\ &
------\\ &
\#\#INSTRUCTIONS:\\ &
1. **Narrative Significance**: Assign a high score if the frame captures pivotal plot developments, character milestones, or key conflicts/resolutions. This measures the frame's impact on the overall story. \\&
2. **Uniqueness and Novelty**: Score highly if the frame introduces new elements or showcases significant alterations in the story or setting. This reflects the frame's contribution to refreshing the narrative. \\&
3. **Action and Dynamics**: Give a high score if the frame depicts crucial actions, events, or is characterized by high energy or movement. This assesses the intensity and momentum conveyed by the frame. \\ \\&
\#\#NOTE:
Keep in mind that the descriptions provided may not fully capture the essence of the corresponding image. Therefore, it's crucial to consider the overall context when determining the importance of the central frame. \\&
Assess its significance not only based on the explicit details given but also in the context of the narrative progression and thematic development.  \\ \hline
Example 1  & Please evaluate the importance score of the central frame \#7 in following 13 frames. Be stingy with scores. \\&
    ------
    \#1: A man is standing on a ramp next to a car. \\&
    \#2: A man is standing on a flatbed truck. \\&
    \#3: A man is standing on a ramp next to a car. \\&
    \#4: A man is standing on a ramp with a blue car on it.\\&
    \#5: A man is standing in front of a crowd of people.\\&
    \#6: A blue shirt with a white collar.\\&
    \#7: A close up of a piece of cloth.\\&
    \#8: A purple wall with a blue stripe.\\&
    \#9: A person's arm with a white shirt on.\\&
    \#10: A person is wearing a purple shirt.\\&
    \#11: A man is holding a rock in his hand.\\&
    \#12: A man is sitting on a chair and holding a car hood.\\&
    \#13: A man is holding a car door open while another man is holding a piece of paper.\\&
    ------\\&

Provide your score where the score is an integer value between 0 and 10, with 10 indicating the highest important frame in a context.\\&
DO NOT PROVIDE ANY OTHER OUTPUT TEXT OR EXPLANARION. Only provide the Python dictionary string.\\ Answer &score: 1 \\ \hline

Example 2  & Please evaluate the importance score of the central frame \#4 in following 7 frames. Be stingy with scores.\\&
------
\#1: A group of people are standing on a roadway near a railroad crossing.\\&
\#2: A group of people are standing on a street corner.\\&
\#3: A group of people are standing on a ramp in the middle of a street.\\&
\#4: A group of people are standing on a road that is blocked off.\\&
\#5: A group of people are standing around a car that is stuck in a puddle.\\&
\#6: A group of people are standing around a car that is on its side.\\&
\#7: A group of people are standing around a car that is on its side.\\&
------\\&

Provide your score where the score is an integer value between 0 and 10, with 10 indicating the highest important frame in a context.\\&
DO NOT PROVIDE ANY OTHER OUTPUT TEXT OR EXPLANARION. Only provide the Python dictionary string.\\ Answer &score: 5 \\ \hline
Example 3  & Please evaluate the importance score of the central frame \#6 in following 11 frames. Be stingy with scores.\\&
------\\&
\#1: A group of people are standing in the middle of a street.\\&
\#2: A group of people are standing in front of a traffic light.\\&
\#3: A group of people are standing on a roadway near a railroad crossing.\\&
\#4: A man is standing on a railroad crossing.\\&
\#5: A man is standing on a railroad crossing.\\&
\#6: A car is driving on a street with a red light.\\&
\#7: A car is driving on a road with a man standing next to a railroad crossing.\\&
\#8: A man is pushing a large metal object in front of a train.\\&
\#9: A man is sitting on a couch in the middle of a street.\\&
\#10: A car is driving through a red light.\\&
\#11: A man is standing on a railroad crossing.\\&
------\\&
Provide your score where the score is an integer value between 0 and 10, with 10 indicating the highest important frame in a context.\\&
\# DO NOT PROVIDE ANY OTHER OUTPUT TEXT OR EXPLANARION. Only provide the Python dictionary string.\\Answer &score: 9
 \\
 \bottomrule
 \caption{Full prompts for LLM $\pi$}
 \label{tab:full_llm_prompt}
\end{longtable}

%% file: main.bbl
\begin{thebibliography}{60}
\providecommand{\natexlab}[1]{#1}
\providecommand{\url}[1]{\texttt{#1}}
\expandafter\ifx\csname urlstyle\endcsname\relax
  \providecommand{\doi}[1]{doi: #1}\else
  \providecommand{\doi}{doi: \begingroup \urlstyle{rm}\Url}\fi

\bibitem[Achiam et~al.(2023)Achiam, Adler, Agarwal, Ahmad, Akkaya, Aleman, Almeida, Altenschmidt, Altman, Anadkat, et~al.]{achiam2023gpt}
Josh Achiam, Steven Adler, Sandhini Agarwal, Lama Ahmad, Ilge Akkaya, Florencia~Leoni Aleman, Diogo Almeida, Janko Altenschmidt, Sam Altman, Shyamal Anadkat, et~al.
\newblock Gpt-4 technical report.
\newblock \emph{arXiv preprint arXiv:2303.08774}, 2023.

\bibitem[Alayrac et~al.(2022)Alayrac, Donahue, Luc, Miech, Barr, Hasson, Lenc, Mensch, Millican, Reynolds, et~al.]{alayrac2022flamingo}
Jean-Baptiste Alayrac, Jeff Donahue, Pauline Luc, Antoine Miech, Iain Barr, Yana Hasson, Karel Lenc, Arthur Mensch, Katherine Millican, Malcolm Reynolds, et~al.
\newblock Flamingo: a visual language model for few-shot learning.
\newblock \emph{Advances in neural information processing systems}, 35:\penalty0 23716--23736, 2022.

\bibitem[Almazrouei et~al.(2023)Almazrouei, Alobeidli, Alshamsi, Cappelli, Cojocaru, Debbah, Goffinet, Hesslow, Launay, Malartic, et~al.]{almazrouei2023falcon}
Ebtesam Almazrouei, Hamza Alobeidli, Abdulaziz Alshamsi, Alessandro Cappelli, Ruxandra Cojocaru, M{\'e}rouane Debbah, {\'E}tienne Goffinet, Daniel Hesslow, Julien Launay, Quentin Malartic, et~al.
\newblock The falcon series of open language models.
\newblock \emph{arXiv preprint arXiv:2311.16867}, 2023.

\bibitem[Anil et~al.(2023)Anil, Dai, Firat, Johnson, Lepikhin, Passos, Shakeri, Taropa, Bailey, Chen, et~al.]{anil2023palm}
Rohan Anil, Andrew~M Dai, Orhan Firat, Melvin Johnson, Dmitry Lepikhin, Alexandre Passos, Siamak Shakeri, Emanuel Taropa, Paige Bailey, Zhifeng Chen, et~al.
\newblock Palm 2 technical report.
\newblock \emph{arXiv preprint arXiv:2305.10403}, 2023.

\bibitem[Apostolidis et~al.(2021)Apostolidis, Balaouras, Mezaris, and Patras]{apostolidis2021combining}
Evlampios Apostolidis, Georgios Balaouras, Vasileios Mezaris, and Ioannis Patras.
\newblock Combining global and local attention with positional encoding for video summarization.
\newblock In \emph{2021 IEEE international symposium on multimedia (ISM)}, pages 226--234. IEEE, 2021.

\bibitem[Argaw et~al.(2024)Argaw, Yoon, Heilbron, Deilamsalehy, Bui, Wang, Dernoncourt, and Chung]{argaw2024scaling}
Dawit~Mureja Argaw, Seunghyun Yoon, Fabian~Caba Heilbron, Hanieh Deilamsalehy, Trung Bui, Zhaowen Wang, Franck Dernoncourt, and Joon~Son Chung.
\newblock Scaling up video summarization pretraining with large language models.
\newblock In \emph{Proceedings of the IEEE/CVF Conference on Computer Vision and Pattern Recognition}, pages 8332--8341, 2024.

\bibitem[Brown(2020)]{brown2020language}
Tom~B Brown.
\newblock Language models are few-shot learners.
\newblock \emph{arXiv preprint arXiv:2005.14165}, 2020.

\bibitem[Chen et~al.(2022)Chen, Guo, Yi, Li, and Elhoseiny]{chen2022visualgpt}
Jun Chen, Han Guo, Kai Yi, Boyang Li, and Mohamed Elhoseiny.
\newblock Visualgpt: Data-efficient adaptation of pretrained language models for image captioning.
\newblock In \emph{Proceedings of the IEEE/CVF Conference on Computer Vision and Pattern Recognition}, pages 18030--18040, 2022.

\bibitem[Chiang et~al.(2023)Chiang, Li, Lin, Sheng, Wu, Zhang, Zheng, Zhuang, Zhuang, Gonzalez, et~al.]{chiang2023vicuna}
Wei-Lin Chiang, Zhuohan Li, Zi Lin, Ying Sheng, Zhanghao Wu, Hao Zhang, Lianmin Zheng, Siyuan Zhuang, Yonghao Zhuang, Joseph~E Gonzalez, et~al.
\newblock Vicuna: An open-source chatbot impressing gpt-4 with 90\%* chatgpt quality.
\newblock \emph{See https://vicuna. lmsys. org (accessed 14 April 2023)}, 2\penalty0 (3):\penalty0 6, 2023.

\bibitem[Driess et~al.(2023)Driess, Xia, Sajjadi, Lynch, Chowdhery, Ichter, Wahid, Tompson, Vuong, Yu, et~al.]{driess2023palm}
Danny Driess, Fei Xia, Mehdi~SM Sajjadi, Corey Lynch, Aakanksha Chowdhery, Brian Ichter, Ayzaan Wahid, Jonathan Tompson, Quan Vuong, Tianhe Yu, et~al.
\newblock Palm-e: An embodied multimodal language model.
\newblock \emph{arXiv preprint arXiv:2303.03378}, 2023.

\bibitem[Fajtl et~al.(2019)Fajtl, Sokeh, Argyriou, Monekosso, and Remagnino]{fajtl2019summarizing}
Jiri Fajtl, Hajar~Sadeghi Sokeh, Vasileios Argyriou, Dorothy Monekosso, and Paolo Remagnino.
\newblock Summarizing videos with attention.
\newblock In \emph{Computer Vision--ACCV 2018 Workshops: 14th Asian Conference on Computer Vision, Perth, Australia, December 2--6, 2018, Revised Selected Papers 14}, pages 39--54. Springer, 2019.

\bibitem[Ghauri et~al.(2021)Ghauri, Hakimov, and Ewerth]{ghauri2021supervised}
Junaid~Ahmed Ghauri, Sherzod Hakimov, and Ralph Ewerth.
\newblock Supervised video summarization via multiple feature sets with parallel attention.
\newblock In \emph{2021 IEEE International Conference on Multimedia and Expo (ICME)}, pages 1--6s. IEEE, 2021.

\bibitem[Gygli et~al.(2014)Gygli, Grabner, Riemenschneider, and Van~Gool]{gygli2014creating}
Michael Gygli, Helmut Grabner, Hayko Riemenschneider, and Luc Van~Gool.
\newblock Creating summaries from user videos.
\newblock In \emph{Computer Vision--ECCV 2014: 13th European Conference, Zurich, Switzerland, September 6-12, 2014, Proceedings, Part VII 13}, pages 505--520. Springer, 2014.

\bibitem[He et~al.(2023)He, Wang, Qiu, Bui, Shrivastava, and Wang]{he2023align}
Bo He, Jun Wang, Jielin Qiu, Trung Bui, Abhinav Shrivastava, and Zhaowen Wang.
\newblock Align and attend: Multimodal summarization with dual contrastive losses.
\newblock In \emph{Proceedings of the IEEE/CVF Conference on Computer Vision and Pattern Recognition}, pages 14867--14878, 2023.

\bibitem[Hussain et~al.(2019)Hussain, Muhammad, Ullah, Cao, Baik, and De~Albuquerque]{hussain2019cloud}
Tanveer Hussain, Khan Muhammad, Amin Ullah, Zehong Cao, Sung~Wook Baik, and Victor Hugo~C De~Albuquerque.
\newblock Cloud-assisted multiview video summarization using cnn and bidirectional lstm.
\newblock \emph{IEEE Transactions on Industrial Informatics}, 16\penalty0 (1):\penalty0 77--86, 2019.

\bibitem[Islam et~al.(2024)Islam, Ho, Yang, Nagarajan, Torresani, and Bertasius]{islam2024video}
Md~Mohaiminul Islam, Ngan Ho, Xitong Yang, Tushar Nagarajan, Lorenzo Torresani, and Gedas Bertasius.
\newblock Video recap: Recursive captioning of hour-long videos.
\newblock In \emph{Proceedings of the IEEE/CVF Conference on Computer Vision and Pattern Recognition}, pages 18198--18208, 2024.

\bibitem[Ji et~al.(2019)Ji, Xiong, Pang, and Li]{ji2019video}
Zhong Ji, Kailin Xiong, Yanwei Pang, and Xuelong Li.
\newblock Video summarization with attention-based encoder--decoder networks.
\newblock \emph{IEEE Transactions on Circuits and Systems for Video Technology}, 30\penalty0 (6):\penalty0 1709--1717, 2019.

\bibitem[Jiang and Mu(2022)]{jiang2022joint}
Hao Jiang and Yadong Mu.
\newblock Joint video summarization and moment localization by cross-task sample transfer.
\newblock In \emph{Proceedings of the IEEE/CVF Conference on Computer Vision and Pattern Recognition}, pages 16388--16398, 2022.

\bibitem[Khandelwal et~al.(2023)Khandelwal, Pavlick, and Sun]{khandelwal2023analyzing}
Apoorv Khandelwal, Ellie Pavlick, and Chen Sun.
\newblock Analyzing modular approaches for visual question decomposition.
\newblock \emph{arXiv preprint arXiv:2311.06411}, 2023.

\bibitem[Khashabi et~al.(2020)Khashabi, Min, Khot, Sabharwal, Tafjord, Clark, and Hajishirzi]{khashabi2020unifiedqa}
Daniel Khashabi, Sewon Min, Tushar Khot, Ashish Sabharwal, Oyvind Tafjord, Peter Clark, and Hannaneh Hajishirzi.
\newblock Unifiedqa: Crossing format boundaries with a single qa system.
\newblock \emph{arXiv preprint arXiv:2005.00700}, 2020.

\bibitem[Kojima et~al.(2022)Kojima, Gu, Reid, Matsuo, and Iwasawa]{kojima2022large}
Takeshi Kojima, Shixiang~Shane Gu, Machel Reid, Yutaka Matsuo, and Yusuke Iwasawa.
\newblock Large language models are zero-shot reasoners.
\newblock \emph{Advances in neural information processing systems}, 35:\penalty0 22199--22213, 2022.

\bibitem[Li et~al.(2023{\natexlab{a}})Li, Ke, Gong, and Drummond]{li2023progressive}
Haopeng Li, Qiuhong Ke, Mingming Gong, and Tom Drummond.
\newblock Progressive video summarization via multimodal self-supervised learning.
\newblock In \emph{Proceedings of the IEEE/CVF winter conference on applications of computer vision}, pages 5584--5593, 2023{\natexlab{a}}.

\bibitem[Li et~al.(2022)Li, Li, Xiong, and Hoi]{li2022blip}
Junnan Li, Dongxu Li, Caiming Xiong, and Steven Hoi.
\newblock Blip: Bootstrapping language-image pre-training for unified vision-language understanding and generation.
\newblock In \emph{International conference on machine learning}, pages 12888--12900. PMLR, 2022.

\bibitem[Li et~al.(2023{\natexlab{b}})Li, Li, Savarese, and Hoi]{li2023blip}
Junnan Li, Dongxu Li, Silvio Savarese, and Steven Hoi.
\newblock Blip-2: Bootstrapping language-image pre-training with frozen image encoders and large language models.
\newblock In \emph{International conference on machine learning}, pages 19730--19742. PMLR, 2023{\natexlab{b}}.

\bibitem[Lin et~al.(2023)Lin, Ye, Zhu, Cui, Ning, Jin, and Yuan]{lin2023video}
Bin Lin, Yang Ye, Bin Zhu, Jiaxi Cui, Munan Ning, Peng Jin, and Li Yuan.
\newblock Video-llava: Learning united visual representation by alignment before projection.
\newblock \emph{arXiv preprint arXiv:2311.10122}, 2023.

\bibitem[Liu et~al.(2024)Liu, Li, Wu, and Lee]{liu2024visual}
Haotian Liu, Chunyuan Li, Qingyang Wu, and Yong~Jae Lee.
\newblock Visual instruction tuning.
\newblock \emph{Advances in neural information processing systems}, 36, 2024.

\bibitem[Loshchilov and Hutter(2017)]{loshchilov2017decoupled}
Ilya Loshchilov and Frank Hutter.
\newblock Decoupled weight decay regularization.
\newblock \emph{arXiv preprint arXiv:1711.05101}, 2017.

\bibitem[Min et~al.(2024)Min, Buch, Nagrani, Cho, and Schmid]{min2024morevqa}
Juhong Min, Shyamal Buch, Arsha Nagrani, Minsu Cho, and Cordelia Schmid.
\newblock Morevqa: Exploring modular reasoning models for video question answering.
\newblock In \emph{Proceedings of the IEEE/CVF Conference on Computer Vision and Pattern Recognition}, pages 13235--13245, 2024.

\bibitem[Narasimhan et~al.(2021)Narasimhan, Rohrbach, and Darrell]{narasimhan2021clip}
Medhini Narasimhan, Anna Rohrbach, and Trevor Darrell.
\newblock Clip-it! language-guided video summarization.
\newblock \emph{Advances in neural information processing systems}, 34:\penalty0 13988--14000, 2021.

\bibitem[Otani et~al.(2019)Otani, Nakashima, Rahtu, and Heikkila]{otani2019rethinking}
Mayu Otani, Yuta Nakashima, Esa Rahtu, and Janne Heikkila.
\newblock Rethinking the evaluation of video summaries.
\newblock In \emph{Proceedings of the IEEE/CVF conference on computer vision and pattern recognition}, pages 7596--7604, 2019.

\bibitem[Potapov et~al.(2014)Potapov, Douze, Harchaoui, and Schmid]{potapov2014category}
Danila Potapov, Matthijs Douze, Zaid Harchaoui, and Cordelia Schmid.
\newblock Category-specific video summarization.
\newblock In \emph{Computer Vision--ECCV 2014: 13th European Conference, Zurich, Switzerland, September 6-12, 2014, Proceedings, Part VI 13}, pages 540--555. Springer, 2014.

\bibitem[Radford et~al.(2019)Radford, Wu, Child, Luan, Amodei, Sutskever, et~al.]{radford2019language}
Alec Radford, Jeffrey Wu, Rewon Child, David Luan, Dario Amodei, Ilya Sutskever, et~al.
\newblock Language models are unsupervised multitask learners.
\newblock \emph{OpenAI blog}, 1\penalty0 (8):\penalty0 9, 2019.

\bibitem[Radford et~al.(2021)Radford, Kim, Hallacy, Ramesh, Goh, Agarwal, Sastry, Askell, Mishkin, Clark, et~al.]{radford2021learning}
Alec Radford, Jong~Wook Kim, Chris Hallacy, Aditya Ramesh, Gabriel Goh, Sandhini Agarwal, Girish Sastry, Amanda Askell, Pamela Mishkin, Jack Clark, et~al.
\newblock Learning transferable visual models from natural language supervision.
\newblock In \emph{International conference on machine learning}, pages 8748--8763. PmLR, 2021.

\bibitem[Reimers and Gurevych(2019)]{reimers2019sentence}
Nils Reimers and Iryna Gurevych.
\newblock Sentence-bert: Sentence embeddings using siamese bert-networks.
\newblock \emph{arXiv preprint arXiv:1908.10084}, 2019.

\bibitem[Saad-Falcon et~al.(2023)Saad-Falcon, Barrow, Siu, Nenkova, Yoon, Rossi, and Dernoncourt]{saad2023pdftriage}
Jon Saad-Falcon, Joe Barrow, Alexa Siu, Ani Nenkova, David~Seunghyun Yoon, Ryan~A Rossi, and Franck Dernoncourt.
\newblock Pdftriage: Question answering over long, structured documents.
\newblock \emph{arXiv preprint arXiv:2309.08872}, 2023.

\bibitem[Son et~al.(2024)Son, Park, and Kim]{son2024csta}
Jaewon Son, Jaehun Park, and Kwangsu Kim.
\newblock Csta: Cnn-based spatiotemporal attention for video summarization.
\newblock In \emph{Proceedings of the IEEE/CVF Conference on Computer Vision and Pattern Recognition}, pages 18847--18856, 2024.

\bibitem[Song et~al.(2024)Song, Chai, Wang, Zhang, Zhou, Wu, Chi, Guo, Ye, Zhang, et~al.]{song2023moviechat}
Enxin Song, Wenhao Chai, Guanhong Wang, Yucheng Zhang, Haoyang Zhou, Feiyang Wu, Haozhe Chi, Xun Guo, Tian Ye, Yanting Zhang, et~al.
\newblock Moviechat: From dense token to sparse memory for long video understanding.
\newblock In \emph{Proceedings of the IEEE/CVF Conference on Computer Vision and Pattern Recognition}, pages 18221--18232, 2024.

\bibitem[Song et~al.(2015)Song, Vallmitjana, Stent, and Jaimes]{song2015tvsum}
Yale Song, Jordi Vallmitjana, Amanda Stent, and Alejandro Jaimes.
\newblock Tvsum: Summarizing web videos using titles.
\newblock In \emph{Proceedings of the IEEE conference on computer vision and pattern recognition}, pages 5179--5187, 2015.

\bibitem[Sul et~al.(2024)Sul, Han, and Lee]{sul2024mr}
Jinhwan Sul, Jihoon Han, and Joonseok Lee.
\newblock Mr. hisum: A large-scale dataset for video highlight detection and summarization.
\newblock \emph{Advances in Neural Information Processing Systems}, 36, 2024.

\bibitem[Taori et~al.(2023)Taori, Gulrajani, Zhang, Dubois, Li, Guestrin, Liang, and Hashimoto]{taori2023alpaca}
Rohan Taori, Ishaan Gulrajani, Tianyi Zhang, Yann Dubois, Xuechen Li, Carlos Guestrin, Percy Liang, and Tatsunori~B Hashimoto.
\newblock Alpaca: A strong, replicable instruction-following model.
\newblock \emph{Stanford Center for Research on Foundation Models. https://crfm. stanford. edu/2023/03/13/alpaca. html}, 3\penalty0 (6):\penalty0 7, 2023.

\bibitem[Team et~al.(2023)Team, Anil, Borgeaud, Wu, Alayrac, Yu, Soricut, Schalkwyk, Dai, Hauth, et~al.]{team2023gemini}
Gemini Team, Rohan Anil, Sebastian Borgeaud, Yonghui Wu, Jean-Baptiste Alayrac, Jiahui Yu, Radu Soricut, Johan Schalkwyk, Andrew~M Dai, Anja Hauth, et~al.
\newblock Gemini: a family of highly capable multimodal models.
\newblock \emph{arXiv preprint arXiv:2312.11805}, 2023.

\bibitem[Terbouche et~al.(2023)Terbouche, Morel, Rodriguez, and Othmani]{terbouche2023multi}
Hacene Terbouche, Maryan Morel, Mariano Rodriguez, and Alice Othmani.
\newblock Multi-annotation attention model for video summarization.
\newblock In \emph{Proceedings of the IEEE/CVF Conference on Computer Vision and Pattern Recognition}, pages 3143--3152, 2023.

\bibitem[Touvron et~al.(2023{\natexlab{a}})Touvron, Lavril, Izacard, Martinet, Lachaux, Lacroix, Rozi{\`e}re, Goyal, Hambro, Azhar, et~al.]{touvron2023llama}
Hugo Touvron, Thibaut Lavril, Gautier Izacard, Xavier Martinet, Marie-Anne Lachaux, Timoth{\'e}e Lacroix, Baptiste Rozi{\`e}re, Naman Goyal, Eric Hambro, Faisal Azhar, et~al.
\newblock Llama: Open and efficient foundation language models.
\newblock \emph{arXiv preprint arXiv:2302.13971}, 2023{\natexlab{a}}.

\bibitem[Touvron et~al.(2023{\natexlab{b}})Touvron, Martin, Stone, Albert, Almahairi, Babaei, Bashlykov, Batra, Bhargava, Bhosale, et~al.]{touvron2023llama2}
Hugo Touvron, Louis Martin, Kevin Stone, Peter Albert, Amjad Almahairi, Yasmine Babaei, Nikolay Bashlykov, Soumya Batra, Prajjwal Bhargava, Shruti Bhosale, et~al.
\newblock Llama 2: Open foundation and fine-tuned chat models.
\newblock \emph{arXiv preprint arXiv:2307.09288}, 2023{\natexlab{b}}.

\bibitem[Vaswani et~al.(2017)Vaswani, Shazeer, Parmar, Uszkoreit, Jones, Gomez, Kaiser, and Polosukhin]{vaswani2017attention}
Ashish Vaswani, Noam Shazeer, Niki Parmar, Jakob Uszkoreit, Llion Jones, Aidan~N Gomez, {\L}ukasz Kaiser, and Illia Polosukhin.
\newblock Attention is all you need.
\newblock \emph{Advances in neural information processing systems}, 30, 2017.

\bibitem[Wang et~al.(2019)Wang, Wang, Wang, Wang, Feng, and Tan]{wang2019stacked}
Junbo Wang, Wei Wang, Zhiyong Wang, Liang Wang, Dagan Feng, and Tieniu Tan.
\newblock Stacked memory network for video summarization.
\newblock In \emph{Proceedings of the 27th ACM international conference on multimedia}, pages 836--844, 2019.

\bibitem[Wang et~al.(2020{\natexlab{a}})Wang, Bai, Long, Hu, Chai, Guan, and Wei]{wang2020query}
Junyan Wang, Yang Bai, Yang Long, Bingzhang Hu, Zhenhua Chai, Yu Guan, and Xiaolin Wei.
\newblock Query twice: Dual mixture attention meta learning for video summarization.
\newblock In \emph{Proceedings of the 28th ACM international conference on multimedia}, pages 4023--4031, 2020{\natexlab{a}}.

\bibitem[Wang et~al.(2020{\natexlab{b}})Wang, Liu, Puri, and Metaxas]{wang2020learning}
Lezi Wang, Dong Liu, Rohit Puri, and Dimitris~N Metaxas.
\newblock Learning trailer moments in full-length movies with co-contrastive attention.
\newblock In \emph{Computer Vision--ECCV 2020: 16th European Conference, Glasgow, UK, August 23--28, 2020, Proceedings, Part XVIII 16}, pages 300--316. Springer, 2020{\natexlab{b}}.

\bibitem[Wei et~al.(2022)Wei, Wang, Schuurmans, Bosma, Xia, Chi, Le, Zhou, et~al.]{wei2022chain}
Jason Wei, Xuezhi Wang, Dale Schuurmans, Maarten Bosma, Fei Xia, Ed Chi, Quoc~V Le, Denny Zhou, et~al.
\newblock Chain-of-thought prompting elicits reasoning in large language models.
\newblock \emph{Advances in neural information processing systems}, 35:\penalty0 24824--24837, 2022.

\bibitem[Xu et~al.(2021)Xu, Wang, Ni, Zhu, Sun, and Wang]{xu2021cross}
Minghao Xu, Hang Wang, Bingbing Ni, Riheng Zhu, Zhenbang Sun, and Changhu Wang.
\newblock Cross-category video highlight detection via set-based learning.
\newblock In \emph{Proceedings of the IEEE/CVF International Conference on Computer Vision}, pages 7970--7979, 2021.

\bibitem[Yuan et~al.(2019)Yuan, Tay, Li, Zhou, and Feng]{yuan2019cycle}
Li Yuan, Francis~EH Tay, Ping Li, Li Zhou, and Jiashi Feng.
\newblock Cycle-sum: Cycle-consistent adversarial lstm networks for unsupervised video summarization.
\newblock In \emph{Proceedings of the AAAI Conference on Artificial Intelligence}, pages 9143--9150, 2019.

\bibitem[Zanella et~al.(2024)Zanella, Menapace, Mancini, Wang, and Ricci]{zanella2024harnessing}
Luca Zanella, Willi Menapace, Massimiliano Mancini, Yiming Wang, and Elisa Ricci.
\newblock Harnessing large language models for training-free video anomaly detection.
\newblock In \emph{Proceedings of the IEEE/CVF Conference on Computer Vision and Pattern Recognition}, pages 18527--18536, 2024.

\bibitem[Zhang et~al.(2023)Zhang, Li, and Bing]{zhang2023video}
Hang Zhang, Xin Li, and Lidong Bing.
\newblock Video-llama: An instruction-tuned audio-visual language model for video understanding.
\newblock \emph{arXiv preprint arXiv:2306.02858}, 2023.

\bibitem[Zhang et~al.(2016)Zhang, Chao, Sha, and Grauman]{zhang2016video}
Ke Zhang, Wei-Lun Chao, Fei Sha, and Kristen Grauman.
\newblock Video summarization with long short-term memory.
\newblock In \emph{Computer Vision--ECCV 2016: 14th European Conference, Amsterdam, The Netherlands, October 11--14, 2016, Proceedings, Part VII 14}, pages 766--782. Springer, 2016.

\bibitem[Zhao et~al.(2018)Zhao, Li, and Lu]{zhao2018hsa}
Bin Zhao, Xuelong Li, and Xiaoqiang Lu.
\newblock Hsa-rnn: Hierarchical structure-adaptive rnn for video summarization.
\newblock In \emph{Proceedings of the IEEE conference on computer vision and pattern recognition}, pages 7405--7414, 2018.

\bibitem[Zhao et~al.(2020)Zhao, Li, and Lu]{zhao2020tth}
Bin Zhao, Xuelong Li, and Xiaoqiang Lu.
\newblock Tth-rnn: Tensor-train hierarchical recurrent neural network for video summarization.
\newblock \emph{IEEE Transactions on Industrial Electronics}, 68\penalty0 (4):\penalty0 3629--3637, 2020.

\bibitem[Zhao et~al.(2021)Zhao, Li, Lu, and Li]{zhao2021reconstructive}
Bin Zhao, Haopeng Li, Xiaoqiang Lu, and Xuelong Li.
\newblock Reconstructive sequence-graph network for video summarization.
\newblock \emph{IEEE Transactions on Pattern Analysis and Machine Intelligence}, 44\penalty0 (5):\penalty0 2793--2801, 2021.

\bibitem[Zhao et~al.(2023)Zhao, Zhang, Wang, Fu, Agarwal, Lee, and Sun]{zhao2023antgpt}
Qi Zhao, Ce Zhang, Shijie Wang, Changcheng Fu, Nakul Agarwal, Kwonjoon Lee, and Chen Sun.
\newblock Antgpt: Can large language models help long-term action anticipation from videos?
\newblock \emph{arXiv preprint arXiv:2307.16368}, 2023.

\bibitem[Zheng et~al.(2024)Zheng, Zhang, Zhang, Ye, Luo, Feng, and Ma]{zheng2024llamafactory}
Yaowei Zheng, Richong Zhang, Junhao Zhang, Yanhan Ye, Zheyan Luo, Zhangchi Feng, and Yongqiang Ma.
\newblock Llamafactory: Unified efficient fine-tuning of 100+ language models.
\newblock \emph{arXiv preprint arXiv:2403.13372}, 2024.

\bibitem[Zhu et~al.(2020)Zhu, Lu, Li, and Zhou]{zhu2020dsnet}
Wencheng Zhu, Jiwen Lu, Jiahao Li, and Jie Zhou.
\newblock Dsnet: A flexible detect-to-summarize network for video summarization.
\newblock \emph{IEEE Transactions on Image Processing}, 30:\penalty0 948--962, 2020.

\end{thebibliography}
